\documentclass[twocolumn]{svjour3}

\usepackage[a4paper, margin=2cm]{geometry}
\usepackage{xcolor}

\smartqed
\usepackage{graphicx}
\usepackage{booktabs}
\usepackage{amssymb}
\usepackage{amsmath}
\usepackage{subfigure}
\usepackage{natbib}
\usepackage[]{siunitx}
\usepackage[ruled,vlined]{algorithm2e}
\usepackage{hyperref}
\RequirePackage{fix-cm}

\newcommand{\vc}[1]     {\ensuremath{\boldsymbol{#1}}}
\newcommand{\vx}        {\vc{x}}
\newcommand{\pars}      {\vc{\beta}}
\newcommand{\rules}      {\vc{r}}
\newcommand{\hyper}      {\vc{\eta}}

\newcommand{\vf}        {\vc{f}}

\journalname{Preprint}

\begin{document}

\title{Rule-based Evolutionary Bayesian Learning}

\titlerunning{Rule-based regression}
\author{Themistoklis Botsas,$^{a}$
	Lachlan R. Mason,$^{a,b}$ Omar K. Matar,$^{a,b}$ Indranil Pan$^{a,b,c^*}$ \\
	\newline
	$^{a}$The Alan Turing Institute \\
	$^{b}$Imperial College London \\
	$^{c}$Newcastle University \\
	}

\institute{ $^*$ Corresponding Author \at
          \email{i.pan11@imperial.ac.uk}           
}
\authorrunning{Botsas, Mason, Matar, Pan}
\date{Received: date / Accepted: date}

\maketitle

\begin{abstract}
In our previous work in \cite{botsas2020rule}, we introduced the rule-based Bayesian Regression, a methodology that leverages two concepts: (i) Bayesian inference, for the general framework and uncertainty quantification and (ii) rule-based systems for the incorporation of expert knowledge and intuition. The resulting method creates a penalty equivalent to a common Bayesian prior, but it also includes information that typically would not be available within a standard Bayesian context. 
In this work, we extend the aforementioned methodology with grammatical evolution, a symbolic genetic programming technique that we utilise for automating the rules' derivation. Our motivation is that grammatical evolution can potentially detect patterns from the data with valuable information, equivalent to that of expert knowledge.
We illustrate the use of the rule-based Evolutionary Bayesian learning technique by applying it to synthetic as well as real data, and examine the results in terms of point predictions and associated uncertainty.
\end{abstract}


\section{Introduction}\label{Introduction}

The issues of inclusion of expert knowledge and opinion into statistical contexts \citep{o2019expert}, and interpretability of generic machine learning models \citep{molnar2020interpretable}, have been widely studied. In our previous work  \cite{botsas2020rule}, we aimed to tackle them for regression problems by introducing the Rule-based Bayesian regression methodology; first, expert opinions were translated into a rule base, i.e. simple IF-THEN statements, where knowledge about associations between inputs and outputs assumes a concrete form. Second, these rules where packaged into a Bayesian context in a manner similar to a standard prior. Finally, machine learning or standard statistical techniques were used as the main likelihood model. The resulting methodology had the ability to introduce expert knowledge into models, which was not typically possible solely from standard Bayesian priors.

In this work, we aim to extend the context described above by automating the first step of the process, i.e. the expert knowledge elicitation. In cases where expert knowledge is limited or non-existent, the rule-based Bayesian context can still be used by employing grammatical evolution \citep{ryan1998grammatical}, a genetic programming technique that uses the notion of a ``grammar'' in order to find simple or more complex associations among the inputs and the output variables. These associations not only provide useful insights into the system, but they can also be used to improve fitting in a practical context, using the rule-based Bayesian methodology.

The rest of the manuscript is organized as follows: In Section~\ref{Methodology} we present the main components of the general rule-based Evolutionary Bayesian learning framework. In Section~\ref{Applications}, we validate and apply our methodology using data from four applications: a simple linear model, a one-dimensional advection simulator, and two real datasets (one regarding carbon emissions, and another focusing on the electrical output of a power plant). In Section \ref{discussion}, we address the shortcomings and complications of the methodology. Finally, in Section \ref{conclusion}, we summarise the main takeaways from our work and discuss possible areas of focus for future research.

\section{Methodology}\label{Methodology}

The methodology comprises different algorithms and techniques. The first essential part is a rule-based system that uses IF–THEN logic-based rules in order to quantify practical knowledge; rule-based systems are described in Section~\ref{Rule-systems}. In Section~\ref{Gram-evol}, we present Grammatical Evolution, a genetic programming technique, and the main innovation for the rule-based Bayesian learning methodology that we introduce in this paper. In Section ~\ref{sec:bayes}, we explain the different ways we pair the rule-based systems (derived from Grammatical evolution) with a conventional Bayesian framework. Finally, these frameworks, along with statistical or machine learning models, are used for regression and classification.

\subsection{Rule-based systems}\label{Rule-systems}

Rule-based systems are useful for introducing additional information (usually derived from domain expertise) into a model, on top of the general model structure and the data. In that sense, they offer similar benefits to Bayesian priors. The issue of using the latter, especially in conjunction with machine learning algorithms, is that Bayesian priors address knowledge about the model's parameters, while expert knowledge can usually refer to associations among inputs and outputs. Rule-based systems, either hand-crafted or derived in some automated technique, on the other hand, can easily describe and facilitate the inclusion of such information into a model.

Our \emph{rule-based} definition includes systems that incorporate knowledge in the form of a rule base $R_k$, which can be expressed as:
\begin{equation}
    \delta_k R_k : \text{ if } A_1^k \oplus A_2^k \oplus \dots \oplus A_{m}^k \text{ then } C_k
\end{equation}
where $\delta_k$ is a dichotomous variable indicating the inclusion of the $k$th rule in the system; $A^k_{i}$, $i \in {1, 2,\dots, m }$, is the value of the $i$th antecedent attribute (cause) in the $k$th rule; $l$ is the number of antecedent attributes used in the $k$th rule; $C_k$ is the consequent (effect) in the $k$th rule; and $\oplus \in \{\lor,\land\}$ represents the set of connectives (OR, AND operations) in the rules.

For our methodology, we include a logical-operator-based (AND, OR) combination of all the rules to give rise to a composite rule base: i.e., $\beta_k = 1,\ \forall\ k$, and we use the quantity:
\begin{equation*}
    R_\text{comp} :=  R_1 \oplus R_2 \oplus \dots \oplus R_n
\end{equation*}
This is a versatile framework that can address rules of different nature. In our context, for example, the antecedent attributes can be functions of one input (e.g. an inequality between an input and a summary statistic), or functions of many inputs (e.g. an equality that compares two or more inputs). Similarly, the antecedent attribute can be a function of the output (e.g. an inequality between the output and a number) or even an equality that describes a full model between inputs and outputs. For more concrete examples see Section~\ref{Applications}.

\subsection{Grammatical evolution}\label{Gram-evol}

Grammatical evolution \citep{ryan1998grammatical} is a genetic programming technique used for automatically generating programs, i.e. sequences of instructions, based on some syntax (a popular application being symbolic regression). It is composed of three separate parts.

The first is a user-specified \emph{grammar}, associated with the program's syntax. This is the part that accounts for all the possible symbolic results that can be derived through combinations of expressions, operations and functions. It allows the user to restrict the search space of all possible functions, and therefore injects some version of domain knowledge. In general, a grammar consists of four components: A non-terminal set $\mathcal{N}$, a terminal set $\mathcal{T}$, a start set $\mathcal{S}$, and a set of production rules $\mathcal{P}$. In our context, the grammar pre-defines the nature of the plausible rules (IF-THEN statements) that the algorithm is allowed to produce and assess. In practice, this refers to the nature of the antecedents and consequents (including different ways of combining expressions for more complex antecedent forms), as well as the associations among them.

The second part is the \emph{cost or fitness function}. It refers to the quality assessment of each proposed program. In practice, it is a quantity that grammatical evolution is trying to minimise in order to retrieve the best possible expression constrained by the grammar. For this work, the choice of the cost function depends on the nature of the rules, which we discuss more on the different applications of the next section. In practice it can vary from something trivial, like minimising the number of points that do not abide with a rule, to more classic cost functions, such as the least square error.

The final part required for the full specification of grammatical evolution is an optimisation algorithm that searches the space defined by the grammar and attempts to find the program that minimises the cost function. Given the symbolic nature of the problem and difficulty with computing gradients for individual symbolic expressions, this needs to be a population based meta-heuristic algorithm, such as Evolutionary Strategy or Genetic Algorithm. For this work, we used the evolutionary strategy as described in \cite{beyer2002evolution}.
For the applications in Section~\ref{Applications} we use grammatical evolution in order to derive a rule-base, which we then use in combination with a Bayesian context, as described in the next section.

\subsection{Bayesian context}\label{sec:bayes}

In a standard Bayesian context, the posterior density is provided by Bayes' theorem:
\begin{equation}\label{Bayes}
    p(\pars | \vx) = \frac{p(\vx | \pars) p(\pars)}{p(\vx)},
\end{equation}
where $\vx$ are the data, $\pars$ are the model parameters, $p(\vx | \pars)$ is the likelihood, and $p(\pars)$ is the prior density. The likelihood describes the data formation, while the prior defines the distributional nature of the parameters before we take data into consideration, and can potentially account for incorporation of expert knowledge. The marginal likelihood, $p(\vx)$, normalises the aforementioned density product in order to make the posterior $p(\pars | \vx)$ a proper density. The latter is the main quantity of interest within a Bayesian context and describes the updated knowledge about the model parameters after the inclusion of both data and expert knowledge.

In some cases, calculating the marginal likelihood analytically is difficult or intractable, so, instead, we employ specialised algorithms, such as Markov Chain Monte Carlo (MCMC), that approximate the posterior density with the help of the proportionality formula of Bayes' theorem:
\begin{equation}\label{Bayes_prop}
    p(\pars | \vx) \propto p(\vx | \pars) p(\pars).
\end{equation}

As we described in \citet{botsas2020rule}, the incorporation of the rule-based systems into the Bayesian context comes by modifying the prior. For the general case Equation \ref{Bayes_prop} becomes:
\begin{equation}\label{Bayes_rules1}
    p(\pars | \vx, \rules) \propto p(\vx | \pars) p(\pars , \rules),
\end{equation}
where $\rules$ is a random variable associated with the rule-base. The extended prior that is the joint distribution $p(\pars, \rules)$ combines the standard knowledge and distributional form associated with the model parameters and the expert information derived from the rule-base. In practice, this yields a framework similar to a conventional Bayesian context with $\pars$ treated as hyperparameters of $\rules$, and, thus, the joint distribution can be computed, by employing the chain rule:
\begin{equation}\label{chain_rule}
    p(\pars , \rules) = p(\rules | \pars) p(\pars),
\end{equation}
and substituting Equation~\eqref{chain_rule} into Equation~\eqref{Bayes_rules1}, which yields:
\begin{equation}\label{eq:rule_regr}
    p(\pars | \vx, \rules) \propto p(\vx | \pars) p(\rules | \pars) p(\pars),
\end{equation}
or, in case we include hyperparameters $\hyper$, that account for the structure of the rules $\rules$ :
\begin{equation}\label{eq:rule_regr_hyper}
    p(\pars | \vx, \rules, \hyper) \propto p(\vx | \pars) p(\rules | \pars, \hyper) p(\pars) p(\hyper).
\end{equation}

\begin{figure*}
\centering
  \includegraphics[width=0.6\textwidth]{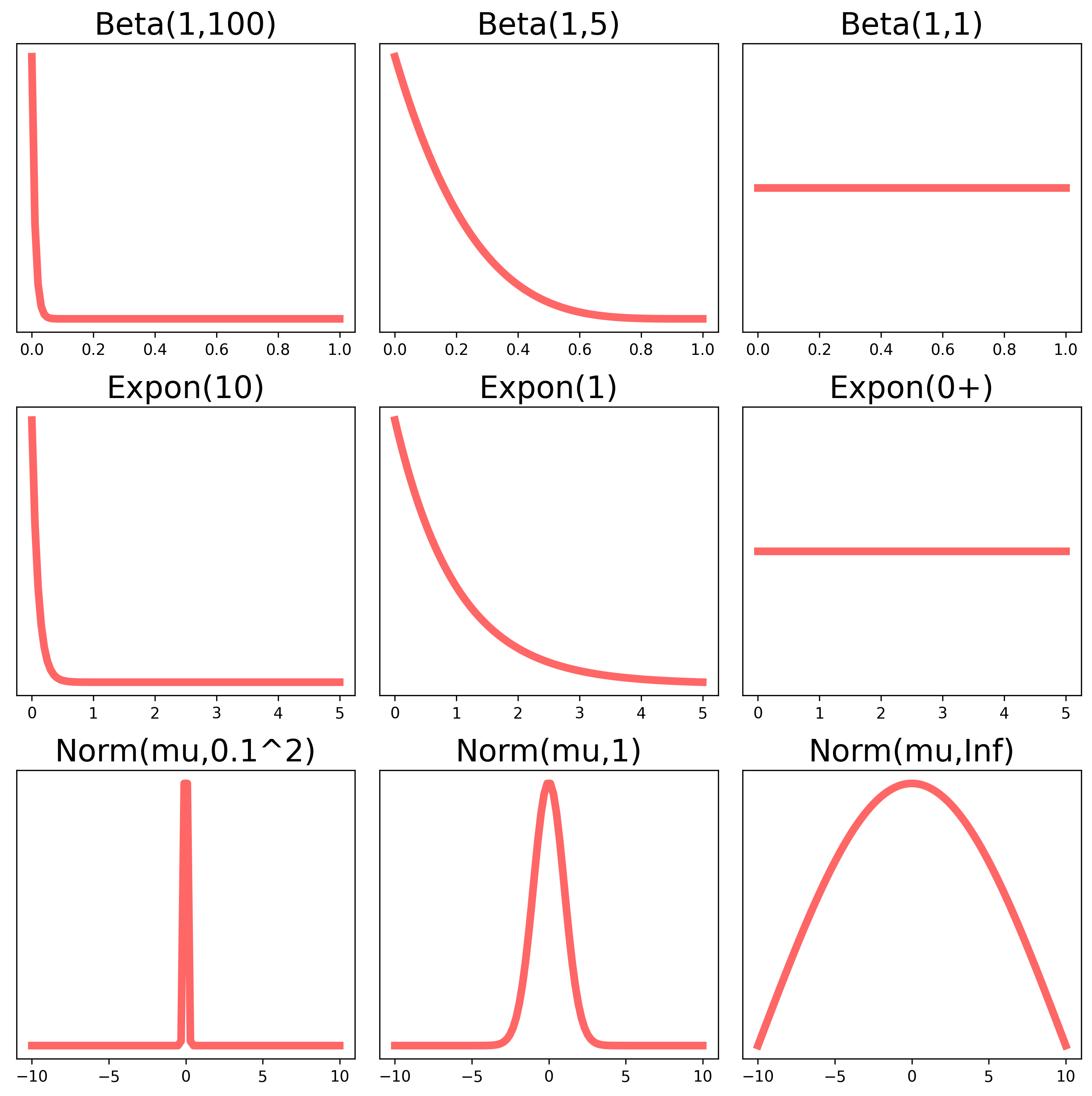}
\caption{Prospective priors for the three variations of the rule distributions. The first row corresponds to the proportion rules (Beta distribution), the second to the total distance (Exponential distribution) and the last one to the piece-wise regression rules (Normal distribution). The first column corresponds to very strict rules, the second to non-strict rules, and the priors in the last column make the rule-based variations equivalent to the standard, non-rule cases. Further explanation of the priors is described in Section~\ref{sec:bayes}.}
\label{fig:distropriors}
\end{figure*}

The term $p(\rules | \pars)$ (or, equivalently, $p(\rules | \pars, \hyper)$) is very general, and can take many different shapes and forms. 
For the purposes of this work, we will examine two main possibilities of useful structures and distributional forms, associated with this quantity.
The first is the one primarily used in \cite{botsas2020rule}: we start by pre-defining discretisations of \textit{rule-input} values, based on the rule-base antecedents. For each proposed set of parameters $\pars$ during the MCMC we compute the number of rule-input values, for which the corresponding outputs violate the respective consequents, and we divide it by the number of all rule-input values. The random variable of this ratio corresponds to $\rules | \pars$. We assign a probability density for this random variable. We use a Beta distribution with parameters $a$ and $b$ ($\operatorname{Beta}(a,b)$) for two reasons. First, $\rules | \pars$ can take values within the interval $[0,1]$ (with $0$ corresponding to no rule-input values violating the respective rule and $1$ to all values violating it). Second, it is very intuitive to incorporate confidence in the rule-base, by adjusting the parameters of the beta distribution, i.e. $\rules | \pars \sim \operatorname{Beta}(1,100)$ corresponds to a strict rule, or `strong confidence' in the rule-base, $\rules | \pars \sim \operatorname{Beta}(1,5)$ to a non-strict rule, and $\rules | \pars \sim \operatorname{Beta}(1,1)$ is equivalent to the non-rule-based approach.

The method described above can yield rules of the form:
\begin{align*}
    R_1&: \text{if} \quad x \leq x_\text{r},\quad \text{then} \quad y \geq y_\text{r},\\
    R_2&: \text{if} \quad x > x_\text{r},\quad \text{then} \quad y < y_\text{r}.
\end{align*}

Two other variations are introduced here. The first is similar to the one described above and depicts the same form of rules, with the main difference being that instead of the \emph{proportion} of rule-input values that violate the rule, we calculate their \emph{total distance} (sum of individual distances) from the rule-boundary. In this case, $\rules | \pars \sim \operatorname{Exp}(\lambda)$, where the choice of the rate parameter $\lambda$ in the Exponential distribution is related to the rule-base confidence: Large values of $\lambda$ correspond to very strict rules and lower values to less strict ones.

The intuition behind the final variation is to consider a penalty by constructing a segmented regressor based on a rule-base, which is independent of the main model used in the likelihood. Initially, we split the data based on the rule breakpoint (i.e. the boundary between the segments) and we compute the \textit{rule-output} values $y'$ for all inputs within each segment. To clarify this, we illustrate the form of rules that this method accommodates:
\begin{align*}
    R_1&: \text{if} \quad x \leq x_\text{r},\quad \text{then} \quad y' = A_1 x + K_1,\\
    R_2&: \text{if} \quad x > x_\text{r},\quad \text{then} \quad y' = A_2 x + K_2.
\end{align*}
For each MCMC sample, $\pars$ is approximated ignoring the $\rules$ dependence. Then $\rules | \pars$ is calculated based on the $\pars$ from the previous step. This procedure repeats for every MCMC iteration, essentially rendering the $\rules | \pars$ equivalent to an \textit{Empirical Bayes prior}.

In practice the vector of rule-outputs $\vc{y'}$ takes the role of $\rules$ and Equation~\eqref{eq:rule_regr} becomes:
\begin{equation}\label{eq:rule_regr_empir_bayes}
    p(\pars | \vx, \vc{y'}) \propto p(\vx | \pars) p(\vc{y'} | \pars) p(\pars),
\end{equation}
where 
\begin{equation}\label{eq:empir_bayes_Gaus}
\vc{y'} | \pars \sim \mathcal{N}(\vf(\vx, \pars^*), \sigma_r).
\end{equation}
For the Equation above, $\vf$ is a function of the data and $\pars^*$. The latter is a point estimate representing the $\pars$ distribution's peak (in a typical Empirical Bayes fashion). The variance $\sigma_r$ is a pre-defined constant associated with the confidence in the rule-base, i.e. a larger variance corresponds to non-strict rules and a smaller variance to strict rules.
In Figure~\ref{fig:distropriors} we indicate some prospective priors for the three variations. 

\section{Applications}\label{Applications}

We now illustrate the use of the methodology by applying it to two synthetic and two real world applications. The first one involves a simple synthetic sub-sample of linear data where we attempt to retrieve the original linear relationship. The second comprises data from a simulator of a one-dimensional advection equation \citep{bar2019learning} where we fit a B-splines model. For the third, we use a multivariate linear regression model in order to fit data that involve CO emissions from gas turbines \citep{kaya2019predicting}. Finally, for the fourth application we use a multivariate logistic regression model to perform classification of the full load electrical power output of a combined cycle power plant \citep{tufekci2014prediction}.

The results are produced with a two-step process; for the first step, we use the gramEvol package from R \citep{noorian2016gramevol} in order to perform the Grammatical evolution optimisation, while, for the second step, the rule-based Bayesian context in produced using the PyMC3 Python package \citep{Salvatier2016}. The source code has been made available online.\footnote{\url{https://github.com/themisbo/Rule-based-Evol-Bayesian-learn}} 

\subsection{Linear regression}\label{sec:lin_reg}

In Section~\ref{sec:lin_reg_data} we illustrate how the synthetic data were produced from a simple linear model. Section~\ref{sec:lin_reg_br} is reserved for the standard Bayesian linear regression analysis. In Section~\ref{sec:lin_reg_ge} we describe how grammatical evolution is used in order to derive a rule-base. Finally, in Section~\ref{sec:lin_reg_rule} we present the analysis for the Bayesian linear regression, which incorporates the rules derived from the previous section.

For the Bayesian analyses, we use a Metropolis–Hastings MCMC \citep{hastings1970monte}, with $2$ chains of \num{120000} iterations each, from which the first \num{20000} are burn-in. For the posterior plots we use thinning of \num{100}. In total, \num{4000} posterior samples are used for the results. For both the intercept and slope priors, we use the same Gaussian distribution $\alpha, \beta  \sim \mathcal{N}(0, 10^2)$, and for the likelihood variance an Exponential distribution $\sigma  \sim {Exp}(1)$.

\subsubsection{Data}\label{sec:lin_reg_data}

\begin{figure}
  \includegraphics[width=\columnwidth]{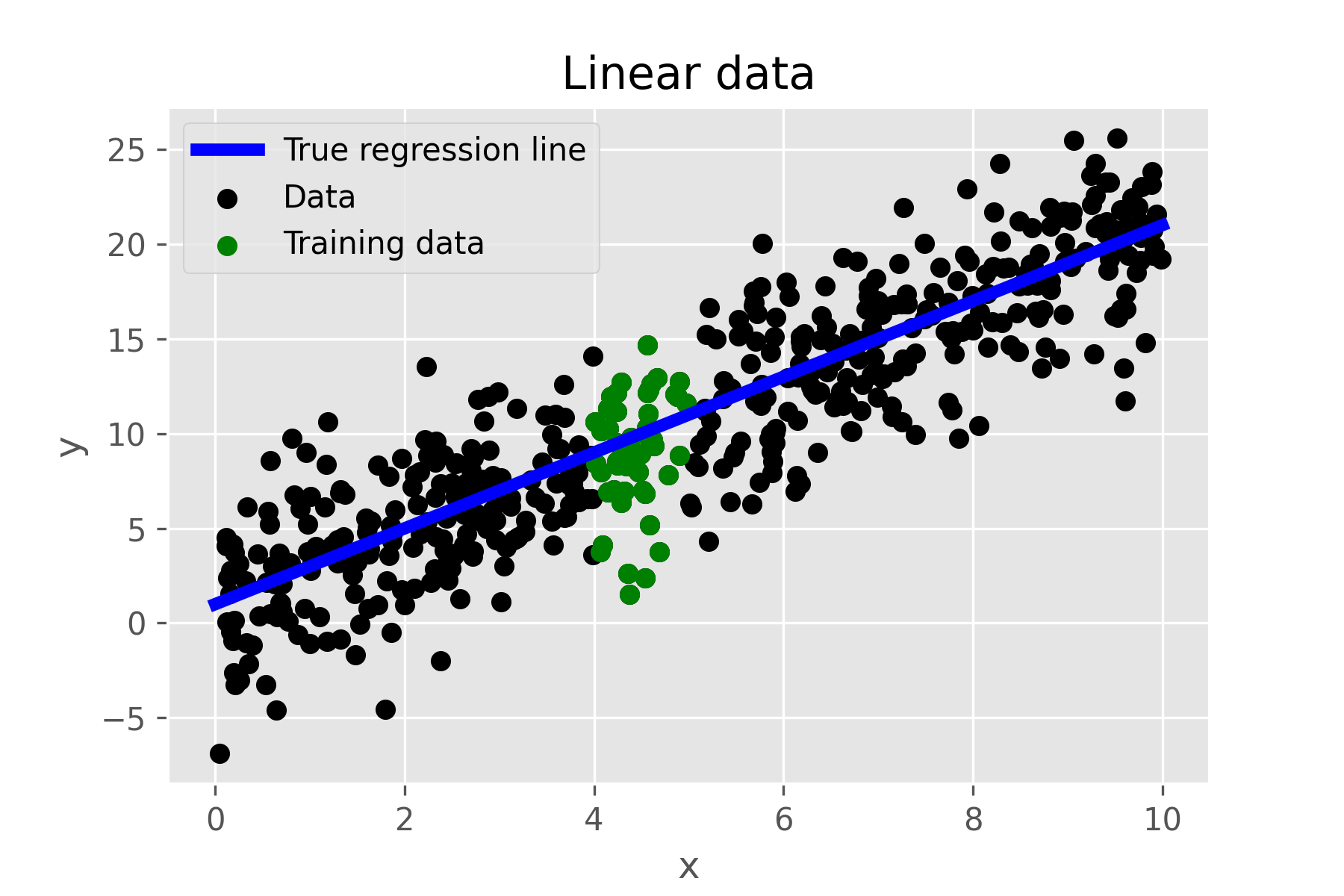}
\caption{Linear regression data.}
\label{fig:linreg_data}
\end{figure}

We produce synthetic linear data, from which we use a small mid-portion as the training set. As we show in the next section this adds a significant amount of uncertainty to the system, and makes recovering the original linear relationship much harder. The goal is to use the grammatical evolution in order to extract the appropriate pattern unprompted, and then use it as additional information through the rule-based Bayesian framework.

We sample $500$ random predictor-values within the interval $[0, 10]$, and we produce the corresponding labels from the \textit{true} regression line $y = 1 + 2x + \epsilon$, where $\epsilon \sim N(0,3^2)$. From those points, we use as training only those within the sub-interval $[4,5]$ which leaves $49$ points for the final analysis. The outcome is shown in Figure~\ref{fig:linreg_data}.

\subsubsection{Bayesian linear regression (BLR)}\label{sec:lin_reg_br}

\begin{table}
\caption{Posterior means $\mu$ and standard deviations $\sigma$ for the parameters of the true values (True), Bayesian linear regression (BLR), proportion rule-based Bayesian linear regression (Prb-BLR), and total distance rule-based Bayesian linear regression (TDrb-BLR).}
\centering
\begin{tabular}{l cc cc}
    \toprule
             & \multicolumn{2}{c}{\textcolor{blue}{True}} & \multicolumn{2}{c}{BLR}\\
    \cmidrule(lr){2-3} \cmidrule(lr){4-5}
    Metric   & \textcolor{blue}{$\mu$}                   & \textcolor{blue}{$\sigma$}           & $\hat{\mu}$                   & $\hat{\sigma}$\\ \midrule
    $\alpha$ & \textcolor{blue}{$1.00$}                  & \textcolor{blue}{$1.73$}                     & $-2.2$                  & $5.74$\\
    $\beta$  & \textcolor{blue}{$2.00$}                  & -                     & $2.53$                  & $1.3$\\ \bottomrule
    \toprule
             & \multicolumn{2}{c}{Prb-BLR}   & \multicolumn{2}{c}{TDrb-BLR} \\
    \cmidrule(lr){2-3} \cmidrule(lr){4-5}
    Metric   & $\hat{\mu}$                   & $\hat{\sigma}$                   & $\hat{\mu}$                      & $\hat{\sigma}$  \\ \midrule
    $\alpha$ & $0.3$                     & $3.44$ & $-1.14$                     & $3.76$  \\
    $\beta$  & $1.96$                     & $077$ & $2.29$                     & $0.85$ \\ \bottomrule
\end{tabular}
\label{tab:linreg_stats}
\end{table}

\begin{table}
\caption{Evaluation metrics for the different linear models: The one without rules (No rules), the one with the proportion rules (Pr. rules) and the one with the total distance rules (T.D. rules).}
\centering
\begin{tabular}{|c|c|c|c|}
\hline

Metric/Model & No rules   & Pr. rules & T.D. rules\\ \hline
        MSE              & \num{48.97}     & \num{33.75} & \num{41.85}\\ \hline
        MAE              & \num{5.71}     & \num{5.09} & \num{5.37}\\ \hline
        WAIC             & \num{249.71}     & \num{249.26} & \num{249.19}\\ \hline
\end{tabular}
\label{tab:linreg_metrics}
\end{table}

\begin{figure}
  \includegraphics[width=\columnwidth]{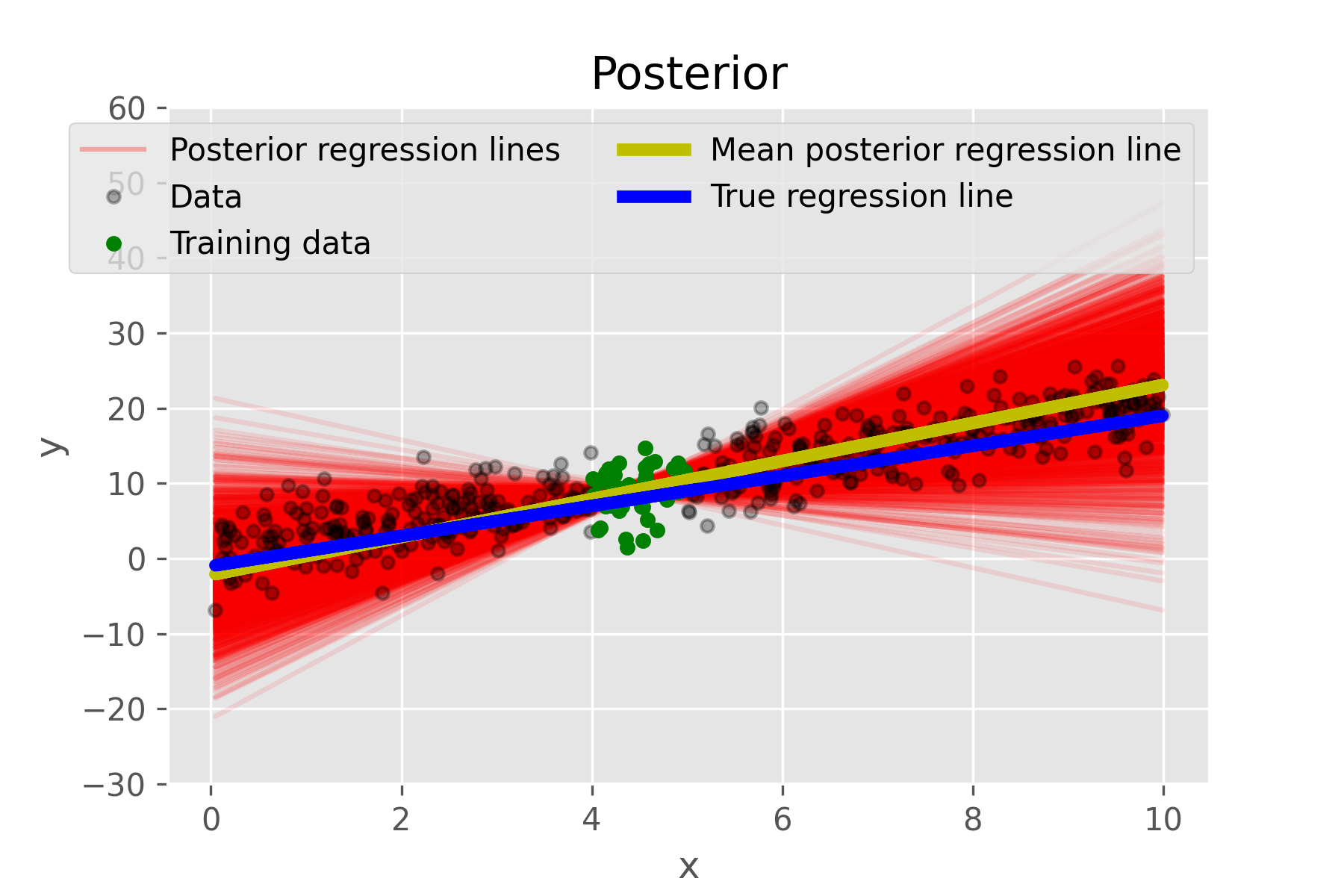}
\caption{Samples from the posterior predictive distribution for Bayesian Linear Regression.}
\label{fig:linreg_post}
\end{figure}

The results for the Bayesian linear regression model are shown in Figure~\ref{fig:linreg_post}. Even though the mean posterior regression line is very close to the true regression line, due to the limited information of the data the uncertainty ranges in the left and right section of the figure (denoted by the red lines) are significantly large. The summary statistics for the MAP (Maximum A Posteriori estimator) parameters are shown in Table~\ref{tab:linreg_stats}, while corresponding metrics are included in Table~\ref{tab:linreg_metrics}.

\subsubsection{Rules derivation (proportion)}\label{sec:lin_reg_ge}

In~\cite{botsas2020rule} we examined how expert knowledge and intuition can be directly translated into a rule-base, and then incorporated into the rule-based Bayesian context. Here on the other hand, we aim to show how the method can still be used, even without information from an expert. For this, we employ the use of grammatical evolution. As explained in Section~\ref{Gram-evol}, we require three components in order to perform grammatical evolution optimisation. We examine each of them individually.

We start with the user-defined grammar and its components in Table~\ref{tab:linear_grammar}.
\begin{table}
\caption{The grammar used for the production of the linear rules.}
\begin{tabular}{l}
$\mathcal{N} = \{expr, comp, x_v, y_v\}$ \\
$\mathcal{T} = \{x,y,>,<,\leq,\geq, 4.1, \dots, 11.0\}$ \\
$\mathcal{S} = <expr>$ \\
\\
$\mathcal{R} = \text{Production rules:} $ \\
$<expr> \quad :== \sum\left(comp(x,x_v)\neq(comp(y,y_v)\right)$ \\
$<comp> \quad :== >|<|\leq|\geq$ \\
$<x_v> \quad :== 4.1|4.15|\dots|4.9 $ \\
$<y_v> \quad :== 7.0|7.05|\dots|11.0$.
\end{tabular}
\label{tab:linear_grammar}
\end{table}
In practice, our goal is to generate different expressions $expr$, which count the points at opposite quadrants. The goal, then, is to minimise $expr$. For this we require the second component of grammatical evolution which is the cost function. Here we use the function $f(x) = x$, which corresponds to direct minimisation of the aforementioned expression.

As mentioned in Section~\ref{Gram-evol}, for the optimisation algorithm required we use the evolution strategy from \cite{beyer2002evolution}. In addition, for all the examples in this work we use the default parameters from the gramEvol package \citep{noorian2016gramevol}: population size of $8$, $25\%$ probability of randomly generated individuals in each generation, mutation chance of $10/(1+\text{population
size})$ and $10,000$ iterations.

The final rules produced are:
\begin{align}\label{rule:312}
    R_1&: \text{if} \quad x < 4.8,\quad \text{then} \quad y \leq 10.65,\\
    R_2&: \text{if} \quad x \geq 4.8,\quad \text{then} \quad  y > 10.65.\nonumber
\end{align}
and the composite rule base ($R_\text{comp}$) is given by
\begin{equation*}
    R_\text{comp} := R_1 \land\ R_2 .
\end{equation*}

\begin{figure}
  \includegraphics[width=\columnwidth]{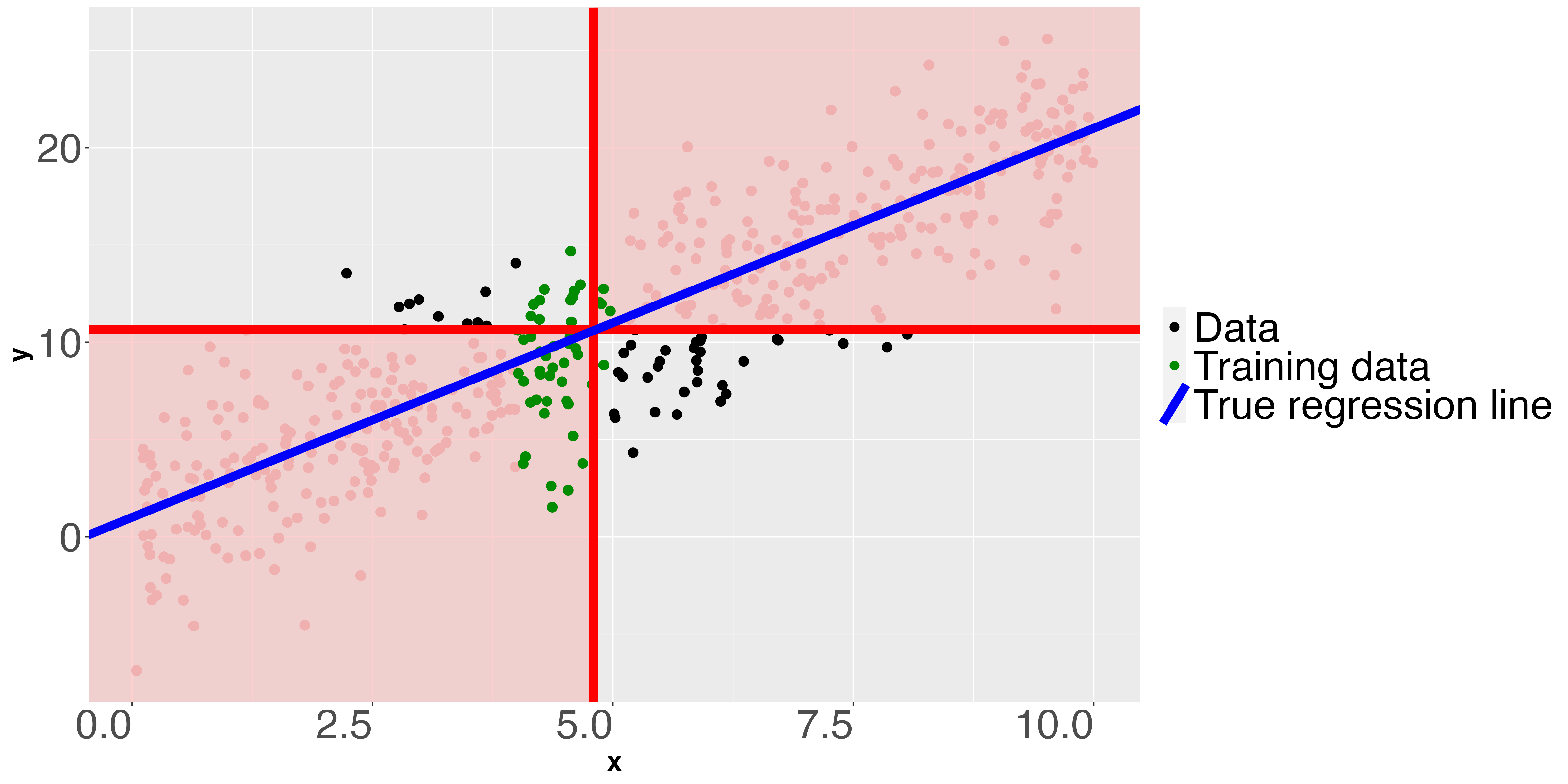}
\caption{Depiction of the rule-base formulated by the Expressions~\ref{rule:312} chosen by grammatical evolution.}
\label{fig:linreg_synth}
\end{figure}

The result is shown graphically in Figure~\ref{fig:linreg_synth}. The values of $x_v$ and $y_v$ chosen by the optimisation algorithm correspond to the thick red vertical and horizontal lines. Given the cost function, the evolution strategy seeks to minimise the points that are not within the red shaded area.

\subsubsection{Rule-based Bayesian linear regression (proportion)}\label{sec:lin_reg_rule}

We use the rules from the previous section for the rule-based Bayesian regression, and also set $\rules | \pars \sim \operatorname{Beta}(1,100)$, which indicates a high level of our confidence in the rules. For the effect of different Beta priors, and therefore different levels of confidence, as well as the variation that includes rule-related hyperparameters see~\cite{botsas2020rule}.

The analytical steps for sampling from the rule-based Bayesian regression model are presented in Algorithm \ref{alg:sampling steps}.

\begin{algorithm}
\SetAlgoLined
Construct discretisations of the rule-input values. Take $n$ equally-spaced points between $n_{min}$ and $n_{mid}$ (antecedent of the first rule) and $n$ equally-spaced points between $n_{mid}$ and $n_{max}$ (antecedent of the second rule)\;
\For{each MCMC iteration}{
Sample new values of $\alpha$ and $\beta$\;
Compute the outputs from the parameter values of step 1 and the discretisations of the rule-input values\;
Calculate the number of these points that violate the corresponding consequents\;
Calculate the ratio of the number of points that violate the rules over the number of all ($2n$) points\;
Compute $\rules | \pars$ (here $\sim \operatorname{Beta}(1,b)$\;
Calculate the un-normalised posterior as the product of the prior, the likelihood and the quantity $\rules | \pars$ from the previous step\;
}
 \caption{Analytical sampling steps}\label{alg:sampling steps}
\end{algorithm}

\begin{figure}
  \includegraphics[width=\columnwidth]{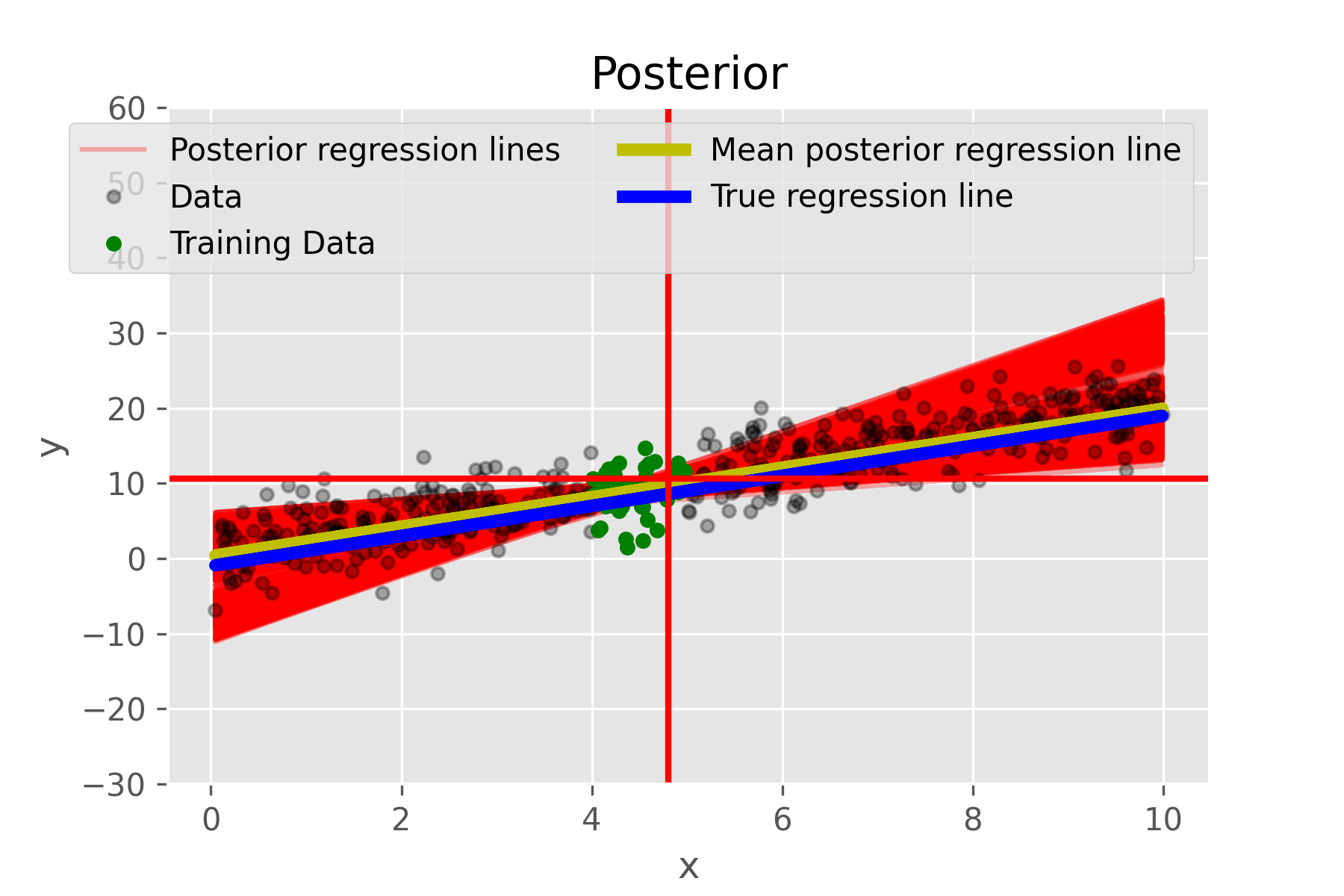}
\caption{Posterior regression lines of the proportion variation of the rule-based Bayesian regression.}
\label{fig:linreg_rulepost}
\end{figure}

The posterior results are shown in Figure~\ref{fig:linreg_rulepost}. We can observe a slight difference for the line that corresponds to the MAP, but the main contrast is the significant reduction in the posteriors' uncertainty (denoted by the red lines). Summary statistics in Table~\ref{tab:linreg_stats} show that the means are slightly closer to their true counterparts for the rule-based Bayesian linear regression, while the variance estimates confirm the uncertainty reduction. Table~\ref{tab:linreg_metrics} presents the mean square error (MSE) and mean absolute error (MAE) for the MAP of the two methods, as well as the Watanabe–Akaike information criterion (WAIC) for the whole chain. We see that the rule-based variation surpasses its conventional counterpart in all metrics.

\subsubsection{Rules derivation (total distance)}\label{sec:lin_reg_ged}

We now examine the same data set for rules based on the total distance, rather than the proportion of the data that violate the rules, as explained in Section~\ref{sec:bayes}. The grammar is similar to the one from Section~\ref{sec:lin_reg_ge}, with the exception of $<expr>$ which becomes:
\begin{multline}\label{grammar:totaldist}
<expr> \quad :== \\ 
ifelse(comp(x, x_v) != comp(y, y_v), y_v, y).
\end{multline}
Given the statement above, $<expr>$ takes the value of $y_v$ for the points where the rule is violated, and the actual point value $y$ otherwise.

The cost function takes the form of the residuals sums of squares (RSS): $f(x) = \sum_{i=1}^n \left( y_i - {expr}_i \right)^2$. Therefore, for the points where the rule is violated, their distance from the rule-boundary is added, and for those that the rule is not violated there is zero increment.

\begin{figure}
  \includegraphics[width=\columnwidth]{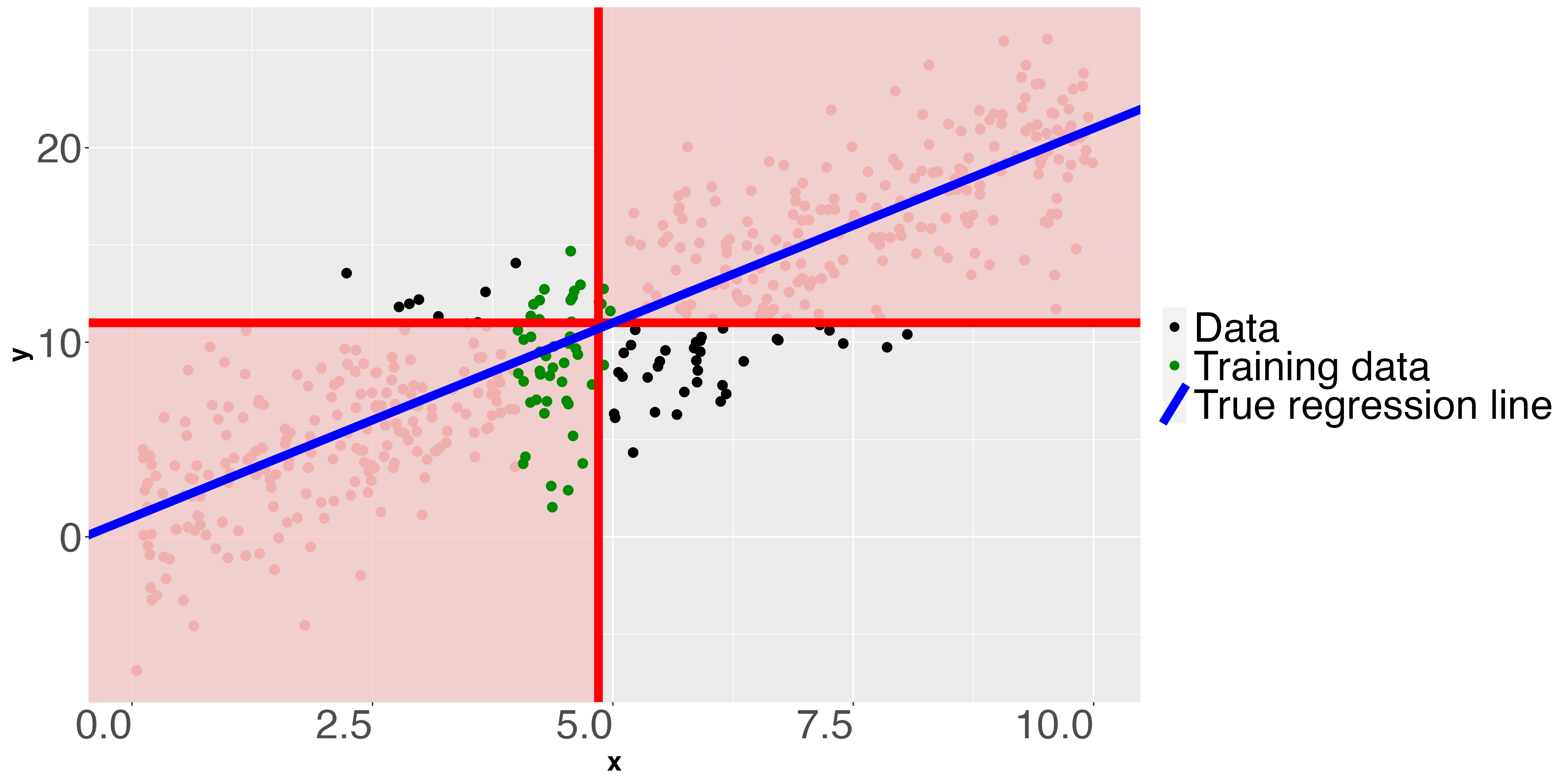}
\caption{Depiction of the rule-base formulated by the Expressions~\ref{rule:314} chosen by grammatical evolution.}
\label{fig:linreg_synthd}
\end{figure}

The rules produced are:
\begin{align}\label{rule:314}
    R_1&: \text{if} \quad x < 4.85,\quad \text{then} \quad y \leq 11,\\
    R_2&: \text{if} \quad x \geq 4.85,\quad \text{then} \quad  y > 11,\nonumber
\end{align}
which are very similar to the corresponding rules from Section~\ref{sec:lin_reg_ge}. This is also confirmed from Figure~\ref{fig:linreg_synthd}.

\subsubsection{Rule-based Bayesian linear regression (total distance)}\label{sec:lin_reg_ruled}

We use the rules from the previous section, and set $\rules | \pars \sim \operatorname{Exp}(10)$, which corresponds to a relatively high level of our confidence.

Sampling from this version of rule-based Bayesian regression is depicted in Algorithm \ref{alg:sampling steps_d}.

\begin{algorithm}
\SetAlgoLined
Construct discretisations of the rule-input values. Take $n$ equally-spaced points between $n_{min}$ and $n_{mid}$ (antecedent of the first rule) and $n$ equally-spaced points between $n_{mid}$ and $n_{max}$ (antecedent of the second rule)\;
\For{each MCMC iteration}{
Sample new values of $\alpha$ and $\beta$\;
Compute the outputs from the parameter values of step 1 and the discretisations of the rule-input values\;
Inspect which of these points violate the corresponding consequents\;
Calculate the total distance (sum of individual euclidean distances) of the points that violate the rules\;
Compute $\rules | \pars$ (here $\sim \operatorname{Exp}(\lambda)$\;
Calculate the un-normalised posterior as the product of the prior, the likelihood and the quantity $\rules | \pars$ from the previous step\;
}
 \caption{Analytical sampling steps}\label{alg:sampling steps_d}
\end{algorithm}

\begin{figure}
  \includegraphics[width=\columnwidth]{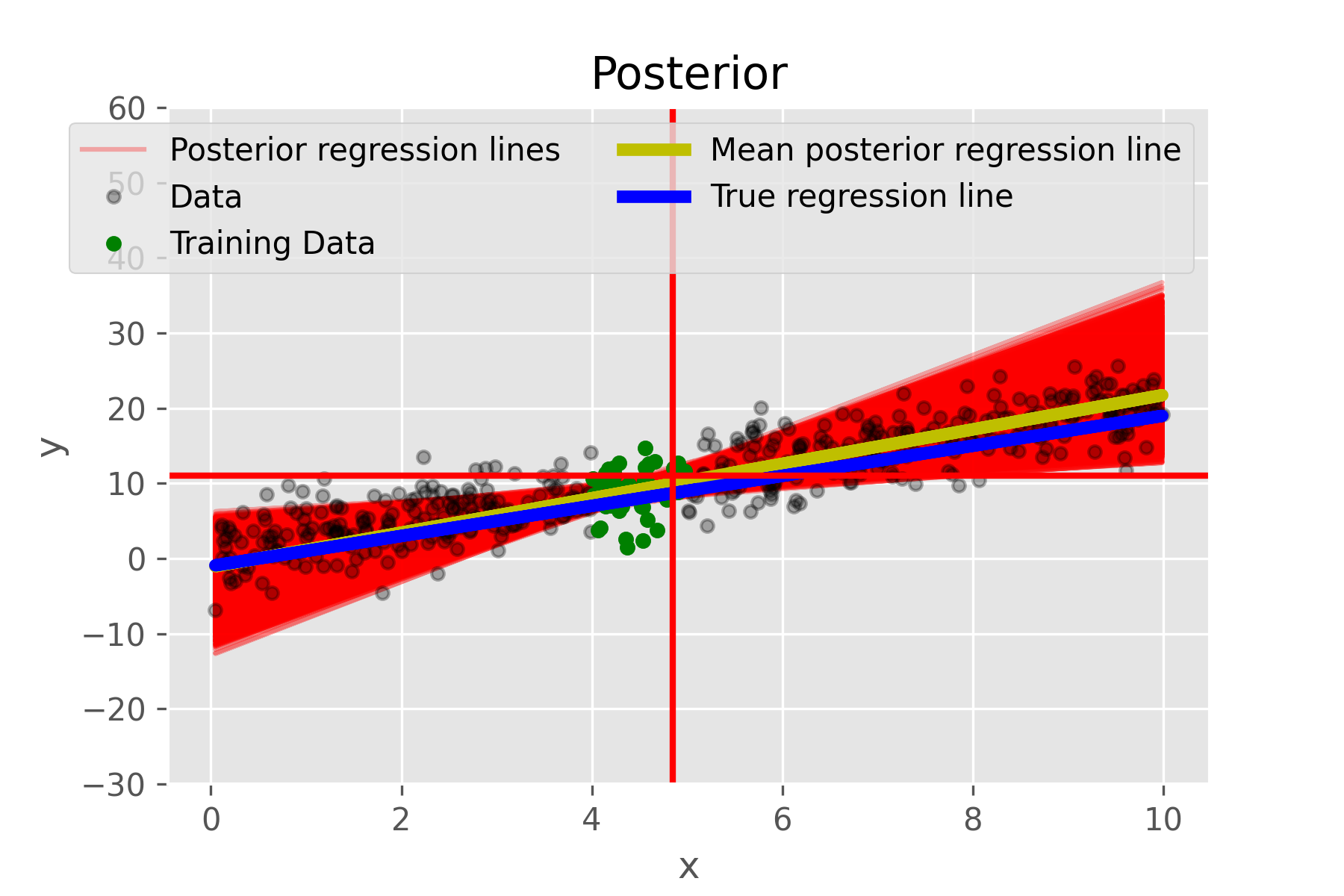}
\caption{Posterior regression lines of the total distance variation of the rule-based Bayesian regressions.}
\label{fig:linreg_rulepostd}
\end{figure}

The posterior plots are shown in Figure~\ref{fig:linreg_rulepostd}. The uncertainty range is somewhere between the wide uncertainty of the case without rules (Figure~\ref{fig:linreg_post}), and the narrow uncertainty of the case with the proportion rules (Figure~\ref{fig:linreg_rulepost}). A larger value of $\lambda$ would move the result towards the latter. The corresponding summary statistics and metrics included in Table~\ref{tab:linreg_stats} and Table~\ref{tab:linreg_metrics} respectively show that the performance of this rule-based Bayesian regression variation is on par with the one from the other rule-based variation, while both rule-based versions perform better than the standard Bayesian regression in terms of all metrics.

\subsubsection{Remarks}

This application outlines the main motivation for this paper; we managed to construct new models that exceed in performance the standard method, by incorporating rules that were automatically derived from the grammatical evolution algorithms, using only grammars, in order to restrict the search space. The algorithms managed to find patterns that were not obvious given the training data and the results were slightly better in terms of the MAP, and significantly better in terms of uncertainty.

\subsection{One-dimensional velocity advection equation}\label{sec:adv_reg}

The velocity advection equation governs transport of momentum by bulk motion. Its one-dimensional form, with a forcing function is
\begin{equation}\label{eq:Burgers1D}
    \frac{\partial u}{\partial t} + u \frac{\partial u}{\partial x} = f(x,t;a,\phi),
\end{equation}
where $u(x,t)$ is the velocity, $x$ the position, $t$ the time, and $f(x,t;a,\phi)$ is the external forcing term with amplitude $a$ and phase $\phi$.

For the Bayesian analyses that follow we fit third-degree B-spline models with $50$ knots. Specifically, we use a reparameterisation that uses the increments of the splines \citep{splines_stan, splines_pymc3} as:
\begin{equation*}
    a_i = a_0 + \sigma_a \sum_{j=1}^{i} \Delta_{aj}.
\end{equation*}
For the priors' specification we use $a_0  \sim \mathcal{N}(0, 0.1^2)$, $\sigma_a  \sim {HalfCauchy}(0.1)$ and $\Delta_{aj}  \sim \mathcal{N}(0, 5^2)$, while the likelihood variance is fixed at $0.002^2$.

For sampling we use the PyMC3 \citep{Salvatier2016} sequential Monte Carlo (SMC) variation, which is a mixture of the Transitional Markov Chain Monte Carlo (TMCMC) \citep{ching2007transitional} and Cascading Adaptive Transitional Metropolis in Parallel (CATMIP) \citep{minson2013bayesian} algorithms. We use \num{10000} draws, which in this implementation also corresponds to the number of chains. Finally, for the posterior plots we use a thinning of $10$.

\subsubsection{Data}\label{sec:adv_reg_data}

\begin{figure}
  \includegraphics[width=\columnwidth]{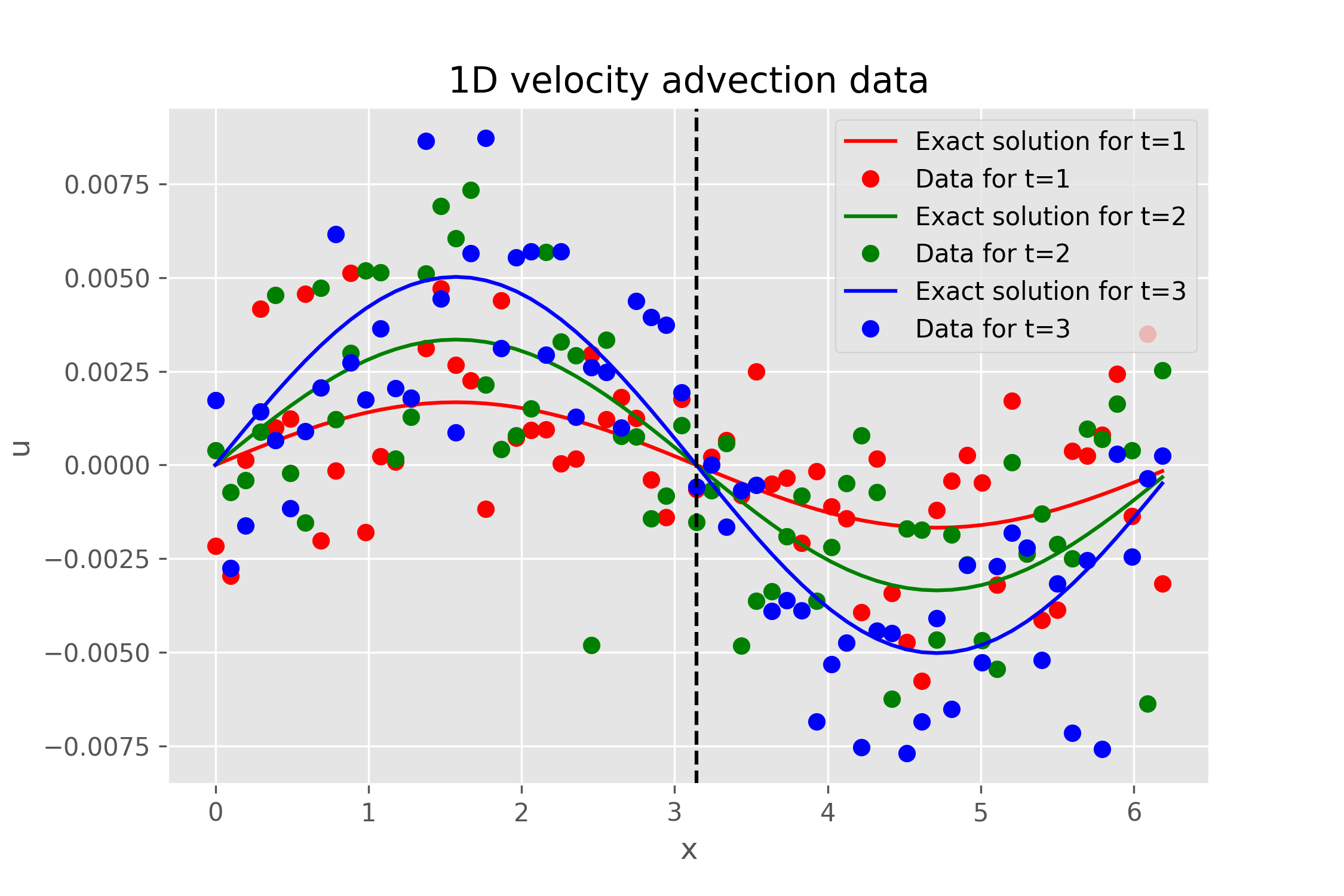}
\caption{One-dimensional velocity advection data.}
\label{fig:1Dadv_data}
\end{figure}

The data for the second application are constructed from a one-dimensional advection velocity equation \citep{bar2019learning} with amplitude $a=0.001$ and phase $\phi = \pi$. We extract the data for three different snapshots $t_j$ (corresponding to $t = 1$, $2$, $3$), before adding a Gaussian error with a standard deviation of $0.002$:
\begin{align*}
    y &= u(x, t) + \epsilon\\
    \epsilon &\sim \mathcal{N}(0,0.002^2).
\end{align*}
The data consist of $96$ points ($32$ values for each snapshot) and are shown in a single plot in Figure~\ref{fig:1Dadv_data} along with the corresponding \emph{true} curves. The point where the curvature changes for all snapshots ($x=\pi$) corresponds to the black dashed line.

\subsubsection{Bayesian B-splines regression}\label{sec:adv_reg_br}

\begin{table}
\caption{Evaluation metrics for the different spline models.}
\centering
\begin{tabular}{|c|c|c|}
\hline

Metric/Model & Without rules   & With rules\\ \hline
        MSE              & \num{3.20e-7}     & \num{2.68e-7}\\ \hline
        MAE              & \num{4.38e-4}     & \num{4.05e-4}\\ \hline
        WAIC              & \num{897.34}     & \num{899.39}\\ \hline
\end{tabular}
\label{tab:1Dadv_metrics}
\end{table}

\begin{figure}
  \includegraphics[width=\columnwidth]{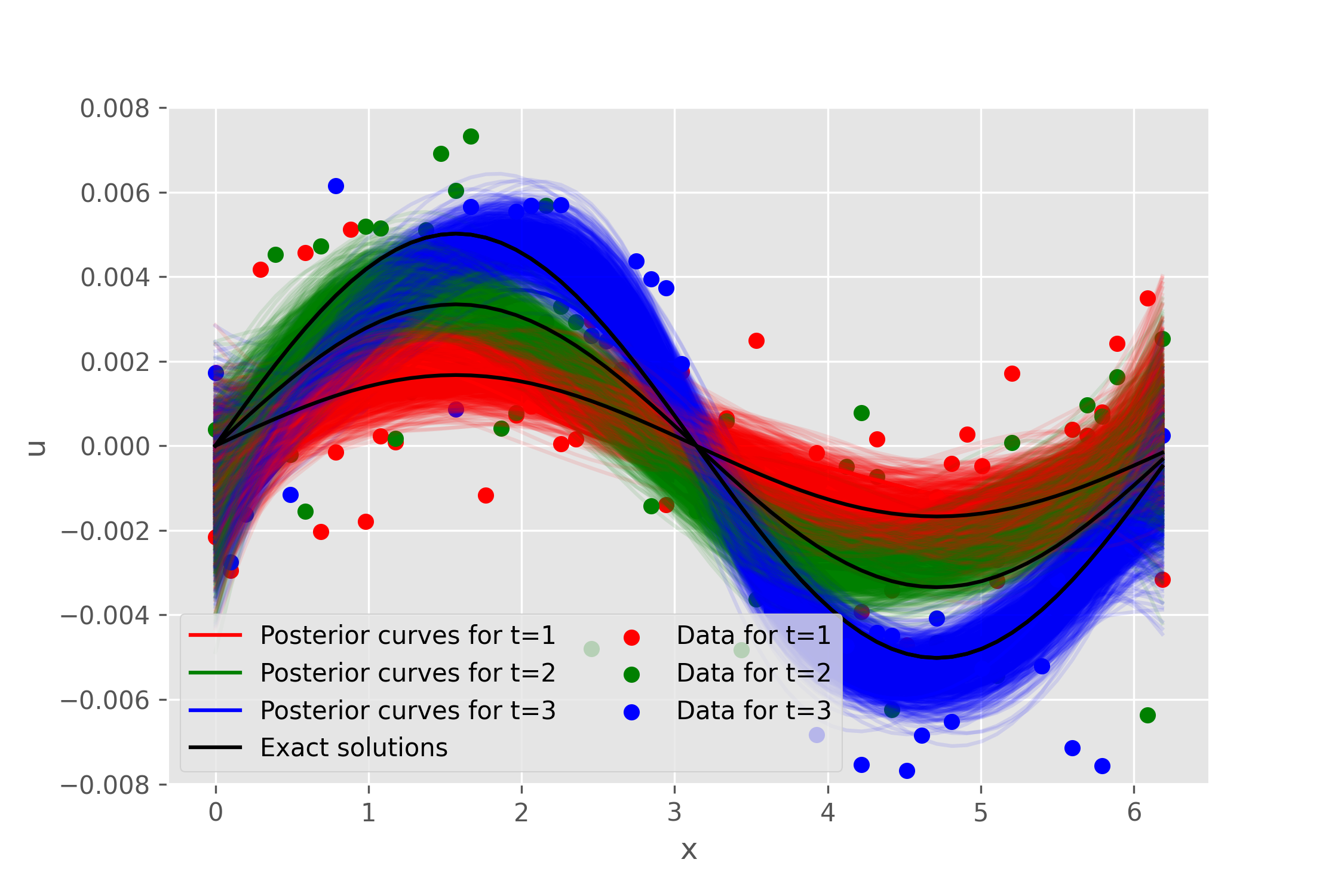}
\caption{Posterior curves for the standard Bayesian regression. The red, green and blue lines are derived from (thinned) samples of the MCMC chain for $t=1,2,$ and $3$ respectively. The black lines denote the exact solutions.}
\label{fig:1Dburgers_rulepost}
\end{figure}

The results of the standard Bayesian B-splines regression are presented in Figure~\ref{fig:1Dburgers_rulepost}. We can observe a lot of overlap among the posterior curves, especially in the left side of the plot, where the curves that correspond to $t=2$ and $t=3$ are clearly flipped, while at the right side of the plot the same curves seem to overlap almost entirely. Similarly in the middle of the plot, the curvature of the curve that corresponds to $t=1$ changes earlier than expected (at $x= \pi$). All these discrepancies are attributed to the Gaussian error included in the data. In the next sections, we examine whether we can use the methodology of this paper in order to derive a better fit even with the limited knowledge that there should be only one point where the curvature of each pair of curves changes.

\subsubsection{Rules derivation}\label{sec:adv_reg_ge}

Once again we specify the requirements for grammatical evolution, starting from the grammar in Table~\ref{tab:1Dadv_grammar}.
\begin{table}
\caption{The grammar used for the production of the one-dimensional velocity advection rules.}
\begin{tabular}{l}
$\mathcal{N} = \{expr, comp, num, u_v\}$ \\
$\mathcal{T} = \{x,u_1,u_2,u_3,>,<, 0.1, \dots, 4.0\}$ \\
$\mathcal{S} = <expr>$ \\
\\
$\mathcal{R} = \text{Production rules:} $ \\
$<expr> \quad :== \sum\left((x < num)\neq(comp(u_v,u_v)\right)$ \\
$<comp> \quad :== >|<$ \\
$<u_v> \quad :== 0.1|0.2|\dots|4.0$.
\end{tabular}
\label{tab:1Dadv_grammar}
\end{table}
The cost function is $f(x) = x$.

This set-up has some major similarities with the one in Section~\ref{sec:lin_reg_ge}, in the sense that we are still attempting to minimise the points in $expr$, while most of the three components remain the same. The only one that changes is the grammar to denote that, instead of counting the points at the quadrants, our goal is to count the points where the outputs that correspond to different time-steps are above (or below) each other.

The rules produced by grammatical evolution are:
\begin{align*}
    R_1&: \text{if} \quad 0 \leq x \leq 2.6,\quad \text{then} \quad u_1 \leq u_2,\\
    R_1'&: \text{if} \quad 2.6 \leq x \leq 2\pi,\quad \text{then} \quad u_1 > u_2,\\
    R_2&: \text{if} \quad 0 \leq x \leq 3.6,\quad \text{then} \quad u_2 \leq u_3,\\
    R_2'&: \text{if} \quad 2.6 \leq x \leq 2\pi,\quad \text{then} \quad u_2 > u_3.
\end{align*}
and the composite rule base ($R_\text{comp}$) is given by
\begin{equation*}
    R_\text{comp} := R_1 \land\ R_1' \land\ R_2 \land\ R_2'.
\end{equation*}

\begin{figure}
  \includegraphics[width=\columnwidth]{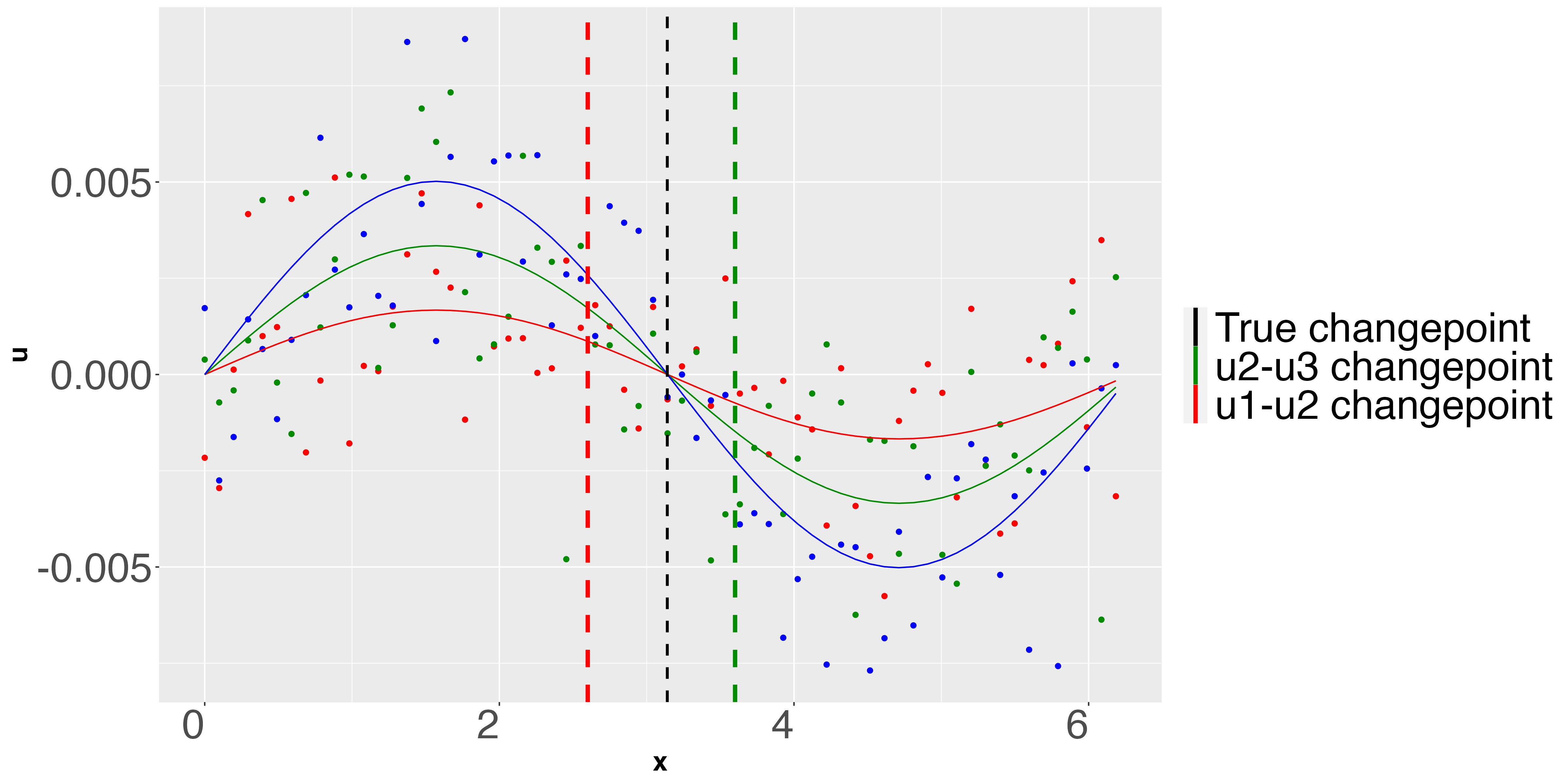}
\caption{Depiction of the rule-base chosen by grammatical evolution.}
\label{fig:1Dadv_synth}
\end{figure}

The result is shown in Figure~\ref{fig:1Dadv_synth}. According to the rules, before the changepoint $x=2.6$, $u_1 \leq u_2$, and after it $u_1 > u_2$. This corresponds to the dashed red line in the plot. Similarly, before the green line changepoint $x=3.6$, $u_2 \leq u_3$, and after it $u_2 > u_3$. It is obvious that grammatical evolution did not manage to produce the optimum rules (which we know from the theory that correspond to the black dashed line $x=\pi$), but the result was close. We contribute this discrepancy to the fact that the data included a fair amount of noise.

\subsubsection{Rule-based Bayesian regression}\label{sec:adv_reg_rule}

\begin{figure}
  \includegraphics[width=\columnwidth]{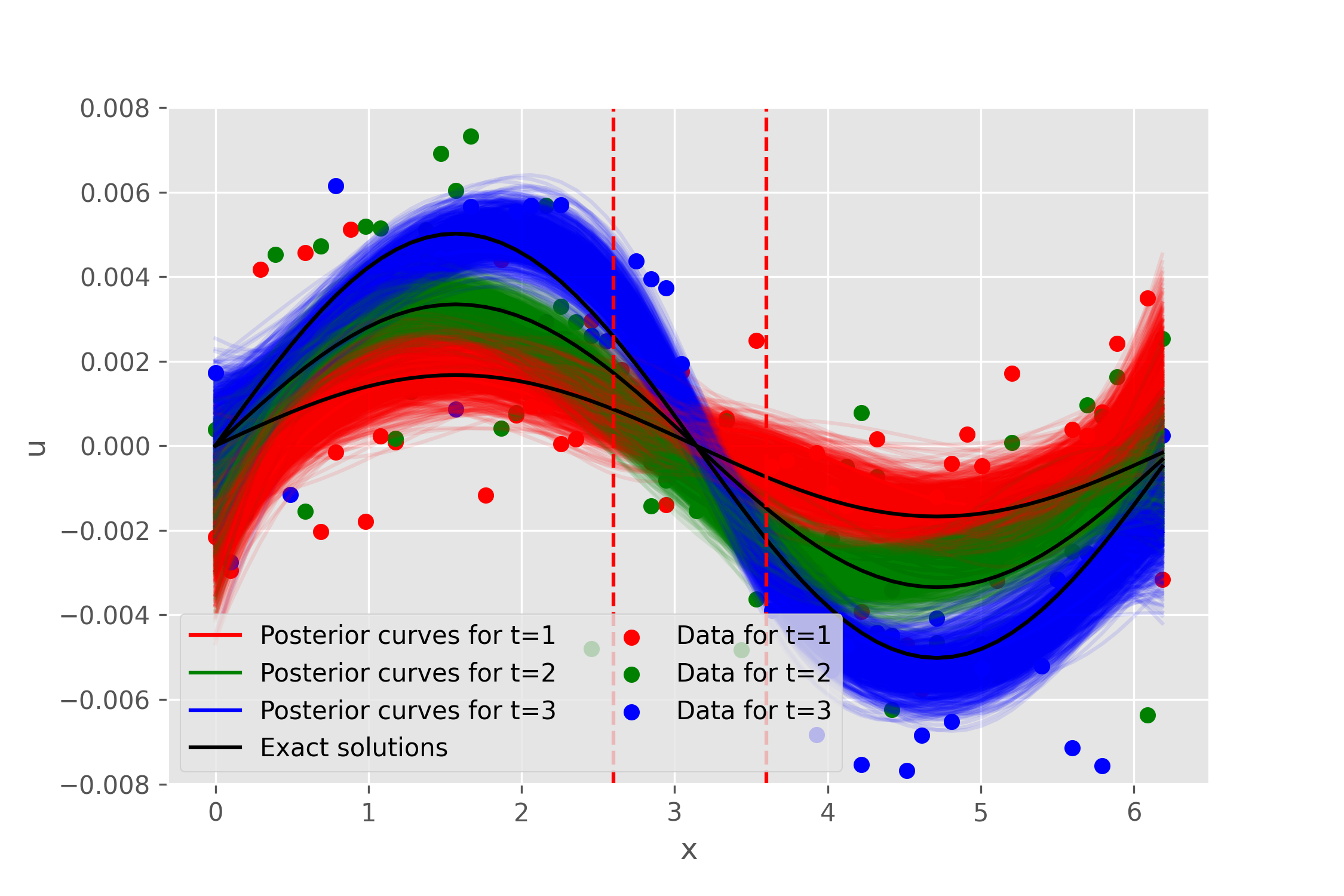}
\caption{Posterior curves for the rule-based Bayesian regression. The red, green and blue lines are (thinned) samples of the MCMC chain for $t=1,2,$ and $3$ respectively. The black lines denote the exact solutions and the red dashed vertical line denotes the rule changepoints.}
\label{fig:1Dadv_rulepost}
\end{figure}

For the rule-based Bayesian analysis we are going to use the same rule conditional distribution as in Section~\ref{sec:lin_reg_rule}, specifically $\rules | \pars \sim \operatorname{Beta}(1,100)$.

The results are shown in Figure~\ref{fig:1Dburgers_rulepost}. The issue at the middle of the plot regarding the early curvature change of the posterior plots that correspond to $t=2$ still remains, but the problems with the left and right edges of the plot have been resolved; the overlap is reduced, the order is correct and the posterior curves are much closer to their true counterparts.

In Table~\ref{tab:1Dadv_metrics} we include relevant metrics. Note that the MSE and MAE are calculated using the MAP and with respect to the true values (the ones that correspond to the curves of the Figures) rather than the observed data (those that correspond to the points of the Figures). Obviously the standard Bayesian regression would yield a better MSE than the rule-based variation if we evaluated the metrics at the observed data points, since, for that case, the MSE is implicitly minimised during training, but our goal here is to try and incorporate any additional knowledge we have in order to derive a result closer to reality. The two metrics mentioned above show that the rule-based Bayesian regression performed better than the non-rule version, which reaffirms the intuition from the Figures. It is interesting to note that the WAIC indicates that the penalty for the point performance increase was additional uncertainty.

\subsubsection{Remarks}

Once again grammatical evolution managed to find useful rules, given the restrictions that we imposed. Even though the rules were not optimal, and more extensive expert knowledge would be beneficial in this case (see Section 4.2 in \cite{botsas2020rule}), there was still a performance increase, which helped to model move towards to the \emph{true} solution.

\subsection{Carbon monoxide (CO) emissions from gas turbines}

For the third application, our aim is to predict $CO$ emission levels of a gas turbine using a multivariate linear regression model. For all the analyses that follow we run a single Metropolis - Hastings chain with \num{100000} draws, in addition to a burn-in of \num{30000} iterations and thinning of \num{100}, which leaves \num{1000} samples for each analysis. Once again we use the PyMC3 package \citep{Salvatier2016}.

\subsubsection{Data}\label{sec:gas_em_data}

\begin{figure}
  \includegraphics[width=\columnwidth]{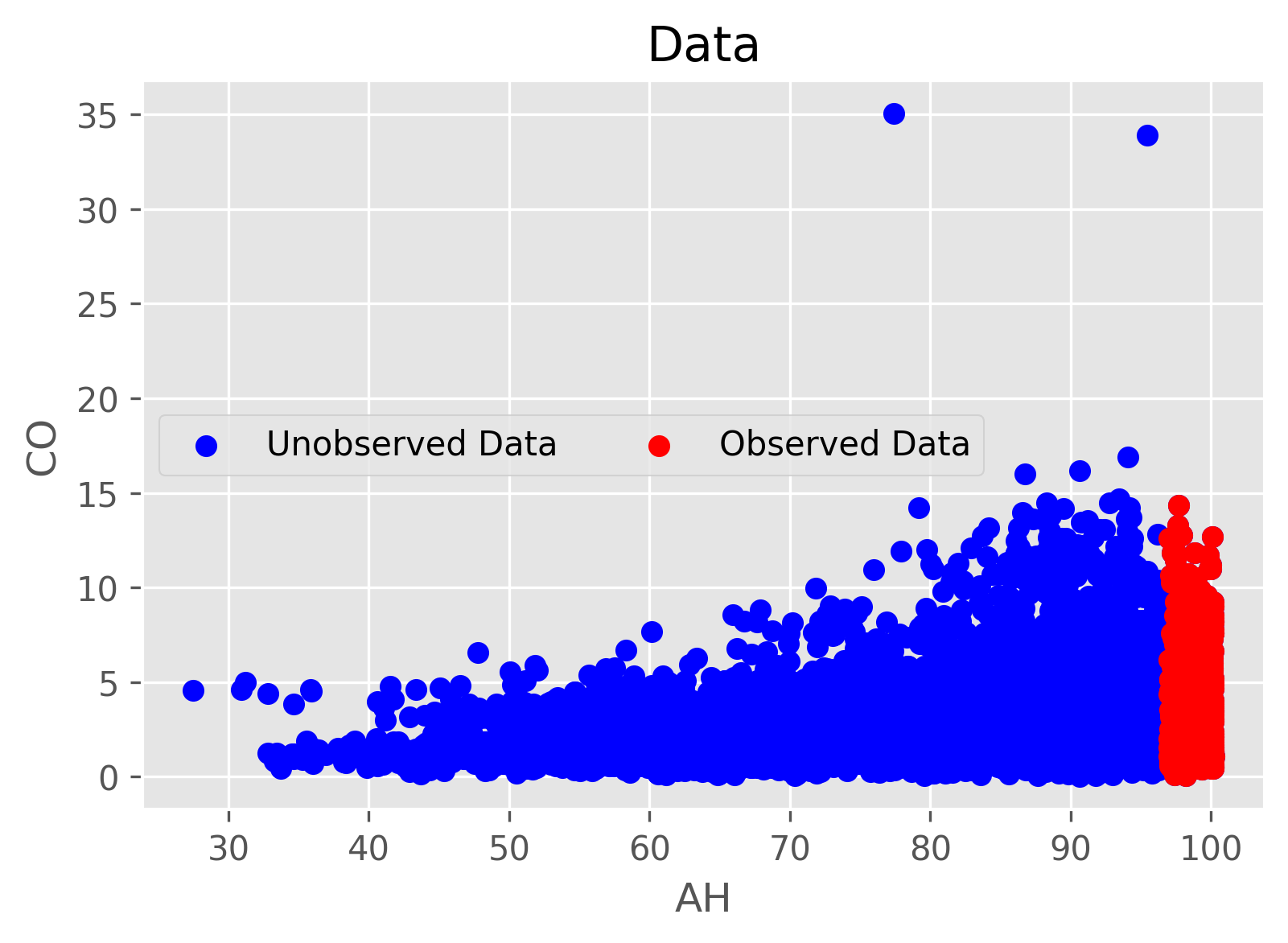}
\caption{Scatterplot of ambient humidity ($AH$) and carbon monoxide emissions ($CO$). The red points show the observed data (used for the model fitting) and the blue points the unobserved data (used for the testing).}
\label{fig:gas_em_data}
\end{figure}

The dataset comes from a field turbine and is described in \cite{kaya2019predicting}. We specifically use the section of the data that correspond to year $2013$. We select four of the features to avoid strong correlations: the ambient temperature $AT$, the ambient humidity $AH$, the air filter difference pressure $AFDP$ and the gas turbine exhaust pressure $GTEP$ and focus on the $CO$ emissions as a single output. Our training set consists of the data where the $AH$ is over $95\%$ of the available data set as shown in Figure~\ref{fig:gas_em_data}, replicating a condition where collection occurs during days with very high humidity, which leaves $547$ data points from the original $7152$. We use the rest of the data to examine how the models generalise. Note that, as we describe in \cite{botsas2020rule}, it is known that there is expert consensus to expect a connection between $AH$ and $CO$.

\subsubsection{Bayesian multivariate linear regression}

For the baseline model we will use multivariate linear regression with parameters the coefficients of all the features. The model is:
\begin{multline*}
    CO = AT_{co}*AT + AH_{co}*AH + AFDP_{co}*AFDP + \\
    GTEP_{co}*GTEP + b + \epsilon,
\end{multline*}
where the carbon monoxide emission level $CO$ is the response, $AT_{co}$, $AH_{co}$, $AFDP_{co}$, and $GTEP_{co}$ are the coefficients of the features that were described in the previous section, $b$ is the intercept and $\epsilon$ is Gaussian error with:
\begin{equation*}
    \epsilon \sim \mathcal{N}(0,\sigma^2).
\end{equation*}

We choose Gaussian distributions for the regression coefficients and intercept, and Exponential for the standard deviation:
\begin{align*}
    AT_{co}, AH_{co}, \dots   &\sim \mathcal{N}(0, 10^2), \\
    b & \sim \mathcal{N}(0, 20^2),\\
    \sigma & \sim \operatorname{Exp}(1).
\end{align*}

\begin{figure}
  \includegraphics[width=\columnwidth]{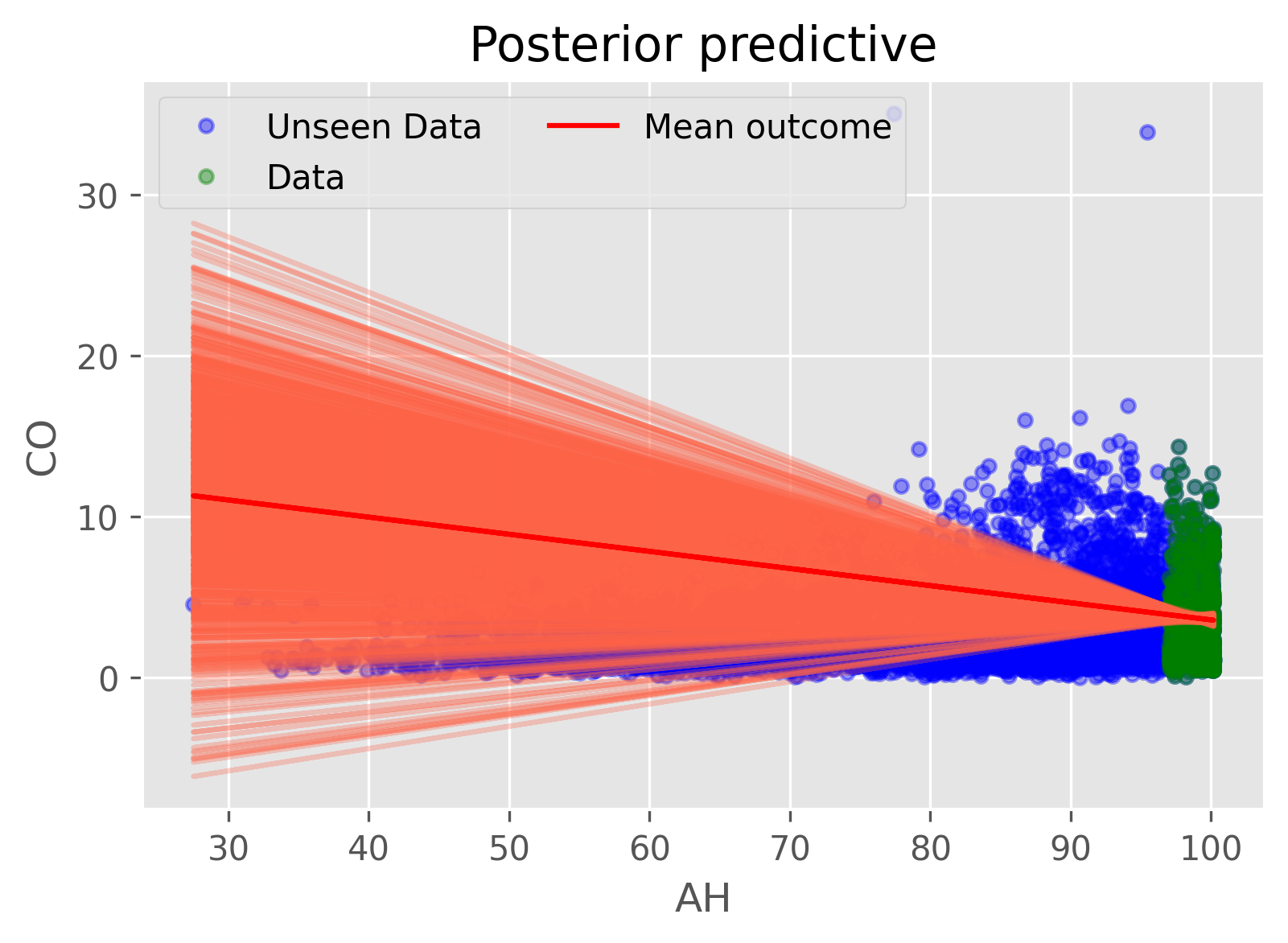}
\caption{Scatterplot of $AH$ and $CO$. The green points are the observed and the blue points the unobserved data. The posterior predictive samples of model without rules are shown in orange and mean in dark red.}
\label{fig:CO_AHnorules}
\end{figure}

\begin{figure}
  \includegraphics[width=\columnwidth]{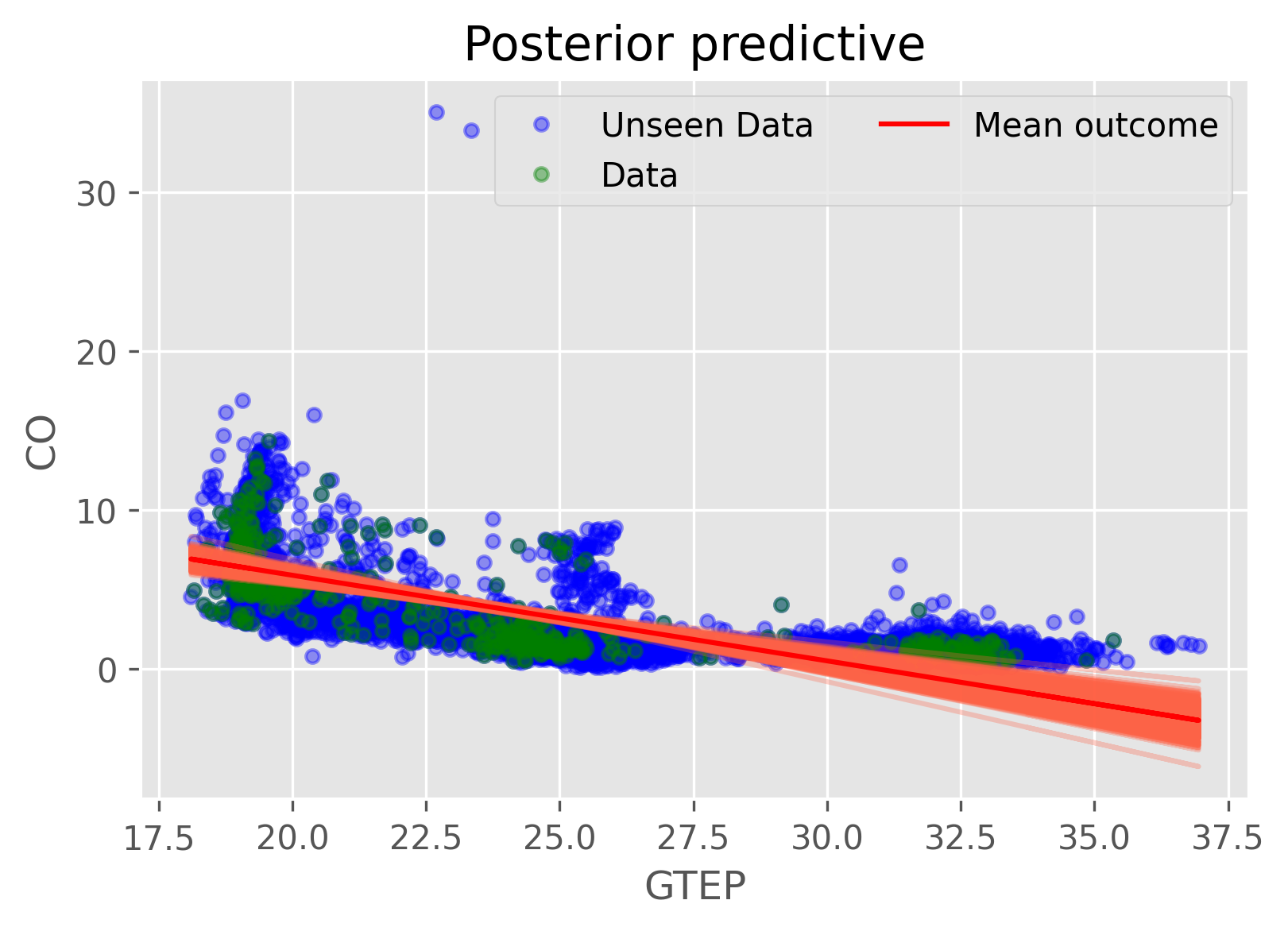}
\caption{Scatterplot of $GTEP$ and $CO$. The green points are the observed and the blue points the unobserved data. The posterior predictive samples of model without rules are shown in orange and mean in dark red.}
\label{fig:CO_GTEPnorules}
\end{figure}

We are going to focus on the $AH - CO$ and $GTEP - CO$ pairs. The corresponding scatterplots, along with the posterior predictive samples are shown in Figure~\ref{fig:CO_AHnorules} and Figure~\ref{fig:CO_GTEPnorules} respectively. In the former we can clearly see that the uncertainty increases drastically away from the training data, while in the latter the slope is slightly different to the one implied by the data.

Evaluation metrics for all the models are included in Table~\ref{emissions_metrics}.

\begin{table}
\caption{Evaluation metrics for the different models.}
\centering
\begin{tabular}{|c||c|c|c|c|}
\hline

Metric/Model & No rules   & $GTEP$ rules            & $AH$ rules\\ \hline
        MSE             & \num{1.36}     & \num{2.19}     & \num{0.83}\\ \hline
        MAE              & \num{0.99}     & \num{1.25}     & \num{0.78}\\ \hline
\end{tabular}
\label{emissions_metrics}
\end{table}

\subsubsection{Rules derivation (piece-wise regression)}\label{rulesderiv-piecewise}

\begin{figure*}
\centering
  \includegraphics[width=0.6\textwidth]{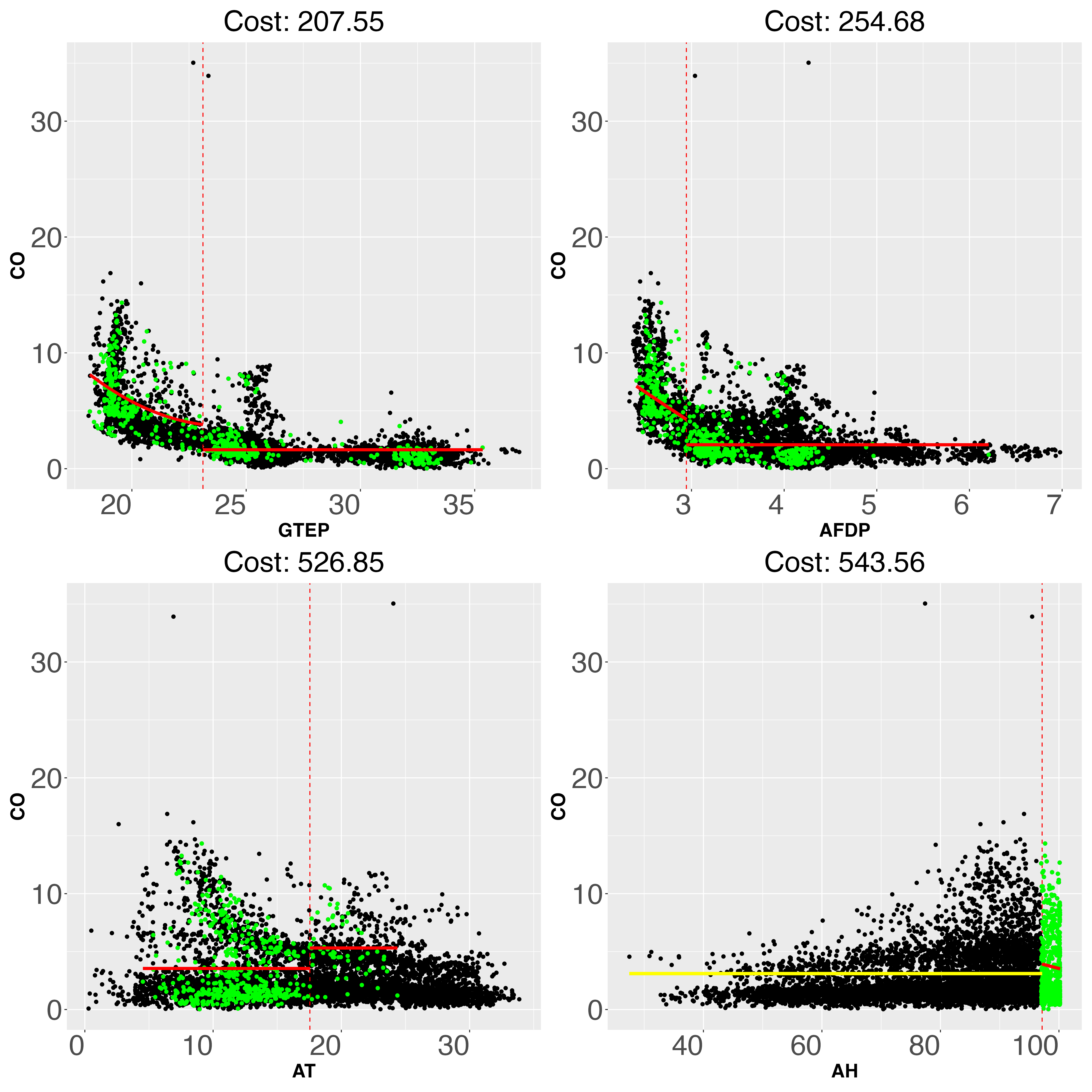}
\caption{Depiction of the rule-bases chosen by grammatical evolution for each feature. The cost on top of each plot indicates the Residual Sum of Squares between the training data (green points) and rule-regression curves (in red). The vertical dotted red lines indicate where the piece-wise models change shape. The yellow line in the last plot is the extension of the rule in the area where there are no training data.}
\label{fig:CO_rulesRegr}
\end{figure*}

The grammar for this case is slightly more complex than the ones in the previous sections. It is presented in Table~\ref{tab:emissionsRegr_grammar}.
\begin{table}
\caption{The grammar used for the production of the carbon monoxide rules.}
\begin{tabular}{l}
$\mathcal{N} = \{op, comp, var, num, num_r, expr, expr_y\}$ \\
$\mathcal{T} = \{AT, AH, \dots,>,<,\leq,\geq, +, -, *, -5.0, \dots, 5.0\}$ \\
$\mathcal{S} = <expr>$ \\
\\
$\mathcal{R} = \text{Production rules:} $ \\
$<expr> \quad :== ifelse((comp(var, num)), expr_y, expr_y )$ \\
$<expr_y> \quad :== op(var,var)|op(var,num_r)|num_r)$ \\
$<op> \quad :== +|-|*$ \\
$<comp> \quad :== >|<|\leq|\geq$ \\
$<var> \quad :== AT|AH|AFDP|GTEP $ \\
$<num_r> \quad :== -5|-4.95|\dots|5 $ \\
$<num> \quad :== -2|-1.95|\dots|2 $.
\end{tabular}
\label{tab:emissionsRegr_grammar}
\end{table}
We need to take into account a few things about this grammar. First, before we feed the data into the algorithm we standardise them. Not only it is going to help with the Bayesian sampling later, but, more importantly, it makes the range $-2$ to $2$ of $<num>$ robust, since it can be used regardless of feature (or response). Second, the algorithm by default samples two different $<expr_y>$, one for when the if statement of $<expr>$ is satisfied and another one when it is not.

We use a residual sum of squares (RSS) type of cost function:
\begin{equation*}
    f(x) = \sum_{i=1}^{N} \left( CO - y' \right)^2,
\end{equation*}

where $N$ is the number of training data and $y'$ are the rule-output values that occurred from $<expr>$ as described in Section~\ref{sec:bayes}.

We attempted to derive rules for all the different features, therefore, we went through all the components of $<var>$ sequentially and run the algorithm again. The results are shown in Figure~\ref{fig:CO_rulesRegr}.

Most of the rules are piece-wise linear, but there are some exceptions. For example the left piece of the $GTEP-CO$ pair is a second degree curve, and the rules associated with the $AT-CO$ pair are both constant. These forms are permitted by the grammar, which, depending on its nature, can allow for more restrictive or more flexible types of rules.

It is very important to note that all these forms are in no way connected to the model that we are trying to fit (in this case a multivariate linear regression model). Regardless of the complexity of the piece-wise models, their only purpose is to add a penalty to the corresponding parameters, and the final model that we will derive in the next section is going to be linear regardless of whether we apply rules or not. The value of the penalty is going to be directly associated with the distance of the actual model (linear) from the rule piece-wise model (which can have various forms).

We are going to focus on the rules associated with the $GTEP-CO$ (best in terms of the cost function) and $AH-CO$ (worst in terms of the cost function) pairs. We remind the reader that $AH$ is the only features for which we have some information (i.e. the training data correspond to the ones with high humidity).

The rules produced by the former pair (before rescaling) are:
\begin{align*}
    R_1&: \text{if} \quad GTEP \leq 23.11,\quad \text{then} \quad y' = GTEP^2,\\
    R_1'&: \text{if} \quad GTEP > 23.11,\quad \text{then} \quad y' = -0.7,
\end{align*}
and for the latter:
\begin{align*}
    R_2&: \text{if} \quad AH \geq 97.16,\quad \text{then} \quad y' = -0.05 AH,\\
    R_2'&: \text{if} \quad AH < 97.16,\quad \text{then} \quad y' = -0.2.
\end{align*}

The composite rule bases ($R_\text{comp}$) are given by
\begin{equation*}
    R_\text{comp} := R_i \land\ R_i', \quad \text{for} \quad i = 1,2.
\end{equation*}

\subsubsection{Rule-based Bayesian multivariate linear regression (piece-wise regression)}

For both pairs we are going to use a Gaussian distribution for $\vc{y'} | \pars$ as described in the last variation of Section~\ref{sec:bayes}. Specifically we use a distribution with mean the residual sum of squares of the rule-output values and the response and standard deviation $0.1$, which indicates relatively high levels of confidence in the rules.

Sampling from the model that incorporates the above is presented in Algorithm~\ref{alg:sampling steps_regr}. 

\begin{algorithm}
\SetAlgoLined
Split the data based on the feature ($i$) and the changepoint ($c$) from the corresponding rule.\;
Calculate the rule-output values from the antecedents of the rules ($y' = f_1(i)$ and $y' = f_2(i)$). \;
\For{each MCMC iteration}{
Sample new values of the intercept and the coefficients of the features\;
Compute $\vc{y'} | \pars$ from the Gaussian with mean the RSS of the rule-output and $CO$ data and standard deviation $sd$: $\vc{y'} | \pars \sim \mathcal{N}(y' - CO,sd^2)$\;
Calculate the un-normalized posterior as the product of the prior, the likelihood and the quantity $\vc{y'} | \pars$ from the previous step\;
}
 \caption{Analytical sampling steps}\label{alg:sampling steps_regr}
\end{algorithm}

\begin{figure}
  \includegraphics[width=\columnwidth]{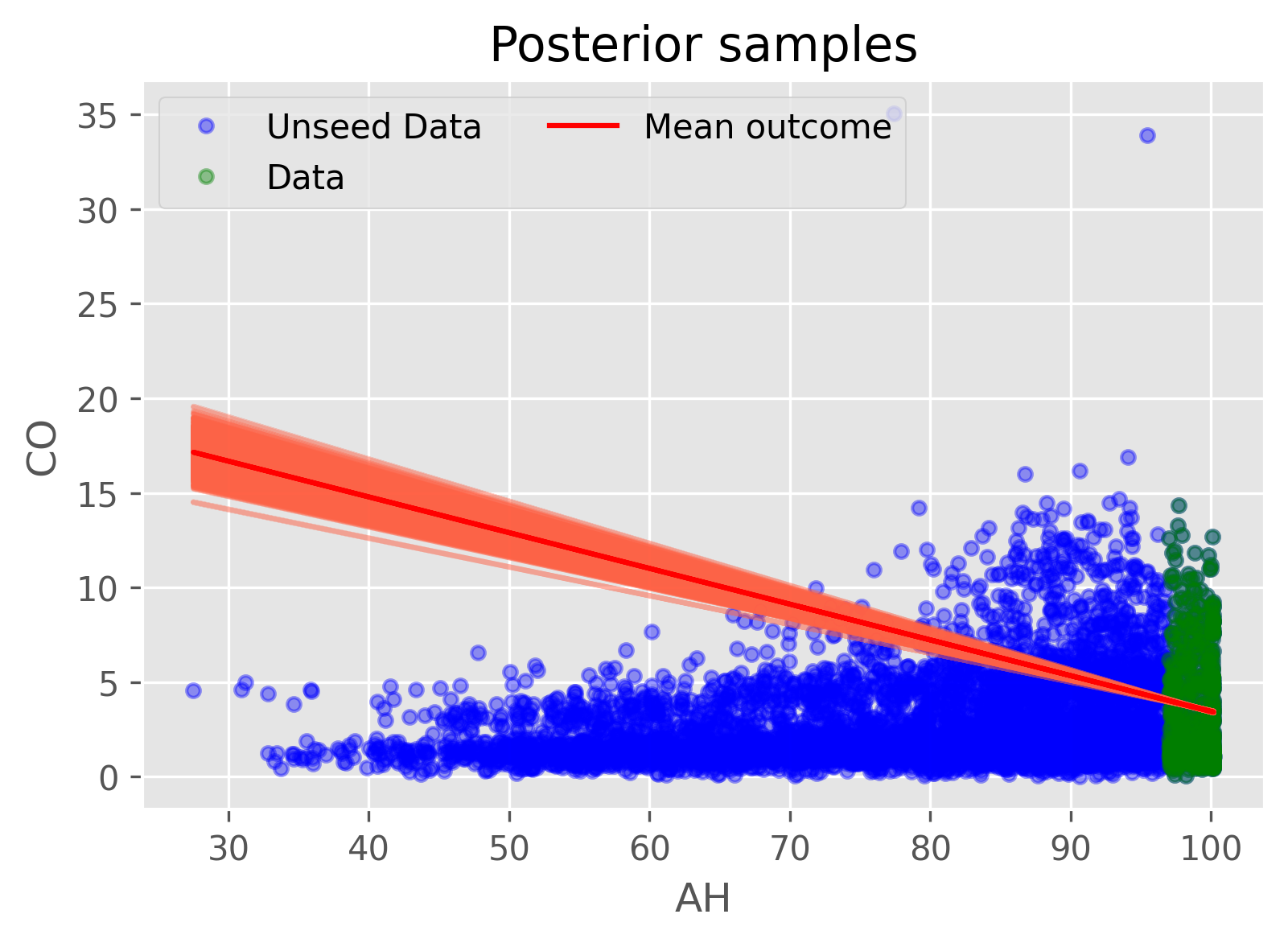}
\caption{Scatterplot of $AH$ and $CO$. The green points are the observed and the blue points the unobserved data. The posterior predictive samples of model without rules are shown in orange and mean in dark red.}
\label{fig:CO_AH_regrGTEPrule01}
\end{figure}

\begin{figure}
  \includegraphics[width=\columnwidth]{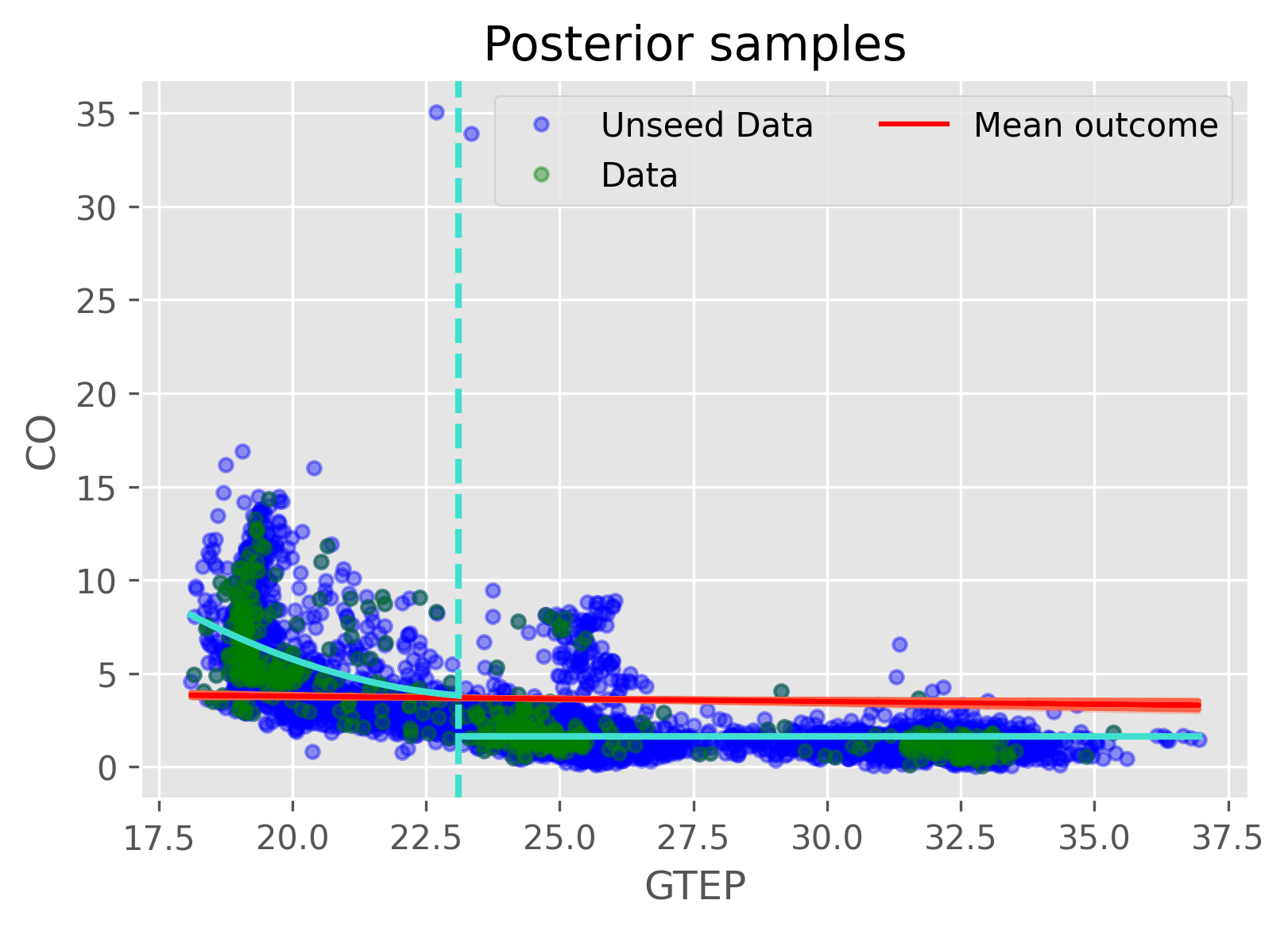}
\caption{Scatterplot of $GTEP$ and $CO$. The green points are the observed and the blue points the unobserved data. The posterior predictive samples of model without rules are shown in orange and mean in dark red. The light blue lines indicate the piece-wise rule model and the vertical dashed light blue line the changepoint.}
\label{fig:CO_GTEP_regrGTEPrule01}
\end{figure}

As explained in the previous section, we are going to run two analyses for the different rules derived by the grammatical evolution. The first one regards the pair $GTEP-CO$; the posterior results for the scatterplots, after we incorporate the corresponding rules, are shown in Figures~\ref{fig:CO_AH_regrGTEPrule01} and \ref{fig:CO_GTEP_regrGTEPrule01}. We can see that in both scatterpolts the uncertainty has decreased substantially. That does not seem to be very helpful for the $AH-CO$ pair, since the the resulting slope is quite different from the one implied by the data.

\begin{figure}
  \includegraphics[width=\columnwidth]{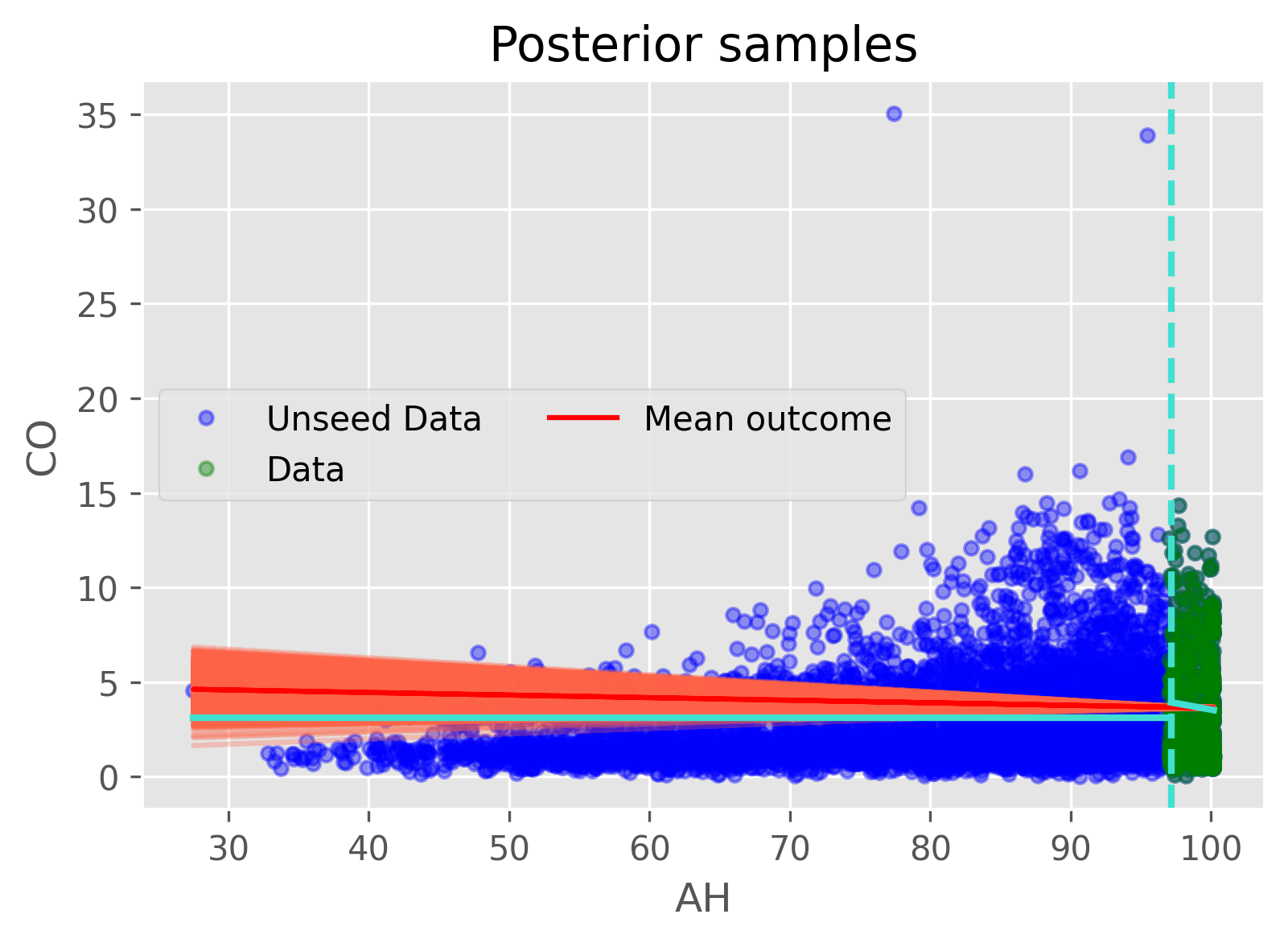}
\caption{Scatterplot of $AH$ and $CO$. The green points are the observed and the blue points the unobserved data. The posterior predictive samples of model without rules are shown in orange and mean in dark red. The light blue lines indicate the piece-wise rule model and the vertical dashed light blue line the changepoint.}
\label{fig:CO_AH_regrAHrule01}
\end{figure}

\begin{figure}
  \includegraphics[width=\columnwidth]{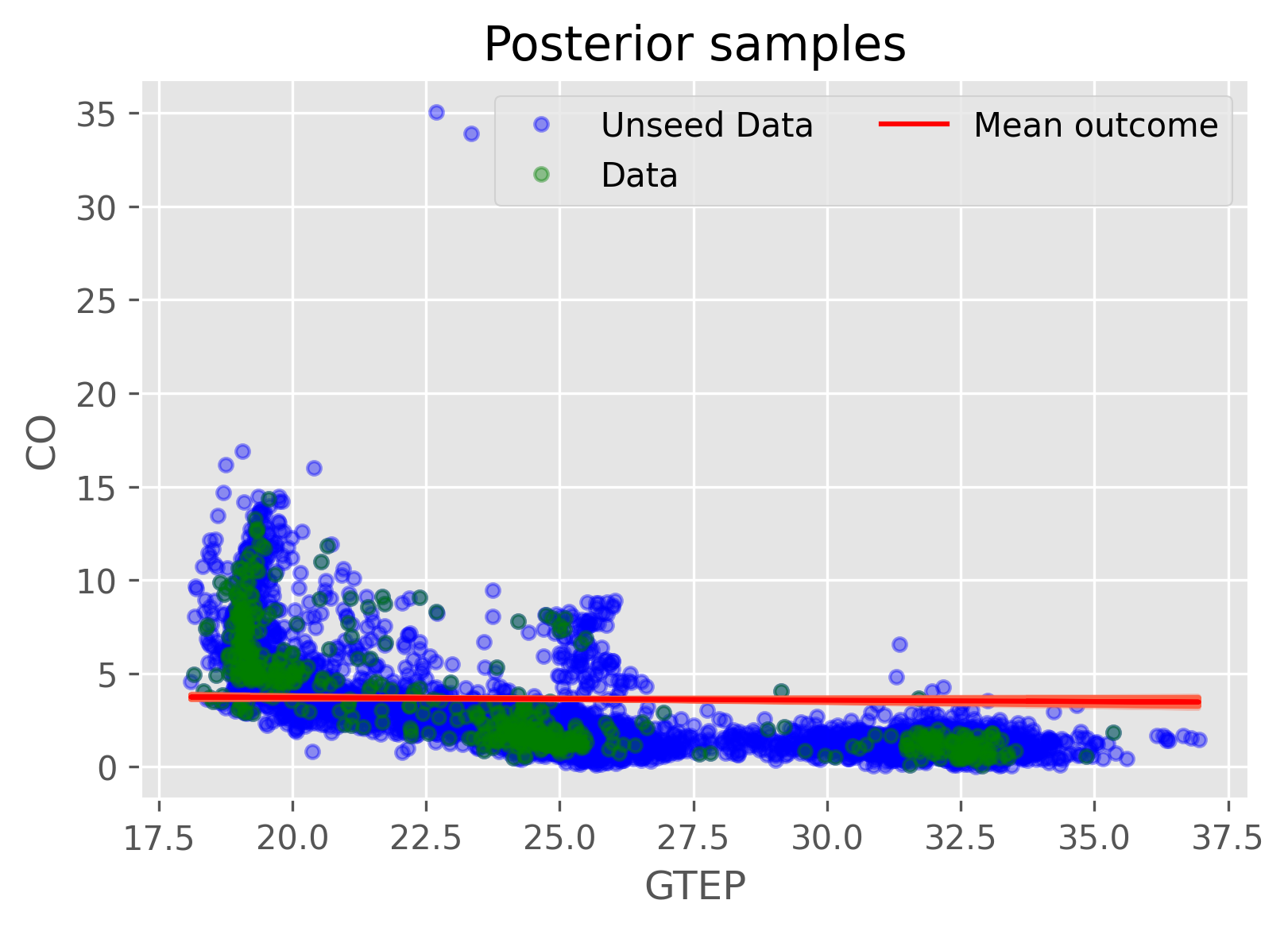}
\caption{Scatterplot of $GTEP$ and $CO$. The green points are the observed and the blue points the unobserved data. The posterior predictive samples of model without rules are shown in orange and mean in dark red.}
\label{fig:CO_GTEP_regrAHrule01}
\end{figure}

The scatterplots for the $AH-CO$ case are presented in Figures~\ref{fig:CO_AH_regrAHrule01} and \ref{fig:CO_GTEP_regrAHrule01}. Once again, the uncertainty has decreased. The slope for the $AH-CO$ pair is quite different from the previous case, approaching the straight line from the left-hand side of the $AH-CO$ rule.

Examining the metrics in Table~\ref{emissions_metrics} we can see that, despite the decrease in the uncertainty, the model with the $GTEP-CO$ rules performed significantly worse than the baseline case, even though it was the best pick from the grammatical evolution algorithm in terms of the cost function. The reason is that, even though the fit is better for the $GTEP-CO$ pair, we should be mindful that the main model is still a multivariate linear regressor, and the fit was poor for some of the other dimensions, thus, making the overall fit poor. On the contrary, the $AH-CO$ fit ended up being significantly better than the baseline fit. Note that $AH$ was the one feature we already had information before the analysis. This is a particularly important result, since it shows the limitations of incorporating the grammatical evolution algorithm into our method, and it also indicates that it works better in combination with domain/expert information.

\subsubsection{Remarks}

In this example we saw how the methodology can be used with real data-sets. The piece-wise regression rules showed that the rule-base selected by grammatical evolution is not always helpful, but, instead, when the algorithm is coupled with additional information and available domain knowledge it can offer significant improvements.

\subsection{Full load electrical power output of a combined cycle power plant}\label{subs:powerplant}

So far we have only dealt with regression problems. For the fourth and final application, our aim is to predict whether the electrical output of a combined cycle power plant (gas and steam turbines) is high or low; therefore we frame it as a classification problem and we tackle it using multivariate logistic regression. For the analyses we run a Metropolis - Hastings chain with \num{100000} draws from the PyMC3 package \citep{Salvatier2016}, in addition to a \num{30000} iterations burn-in and thinning of \num{100}. In total there are \num{1000} final samples for the analysis.

\subsubsection{Data}\label{sec:pe_data}

\begin{figure}
  \includegraphics[width=\columnwidth]{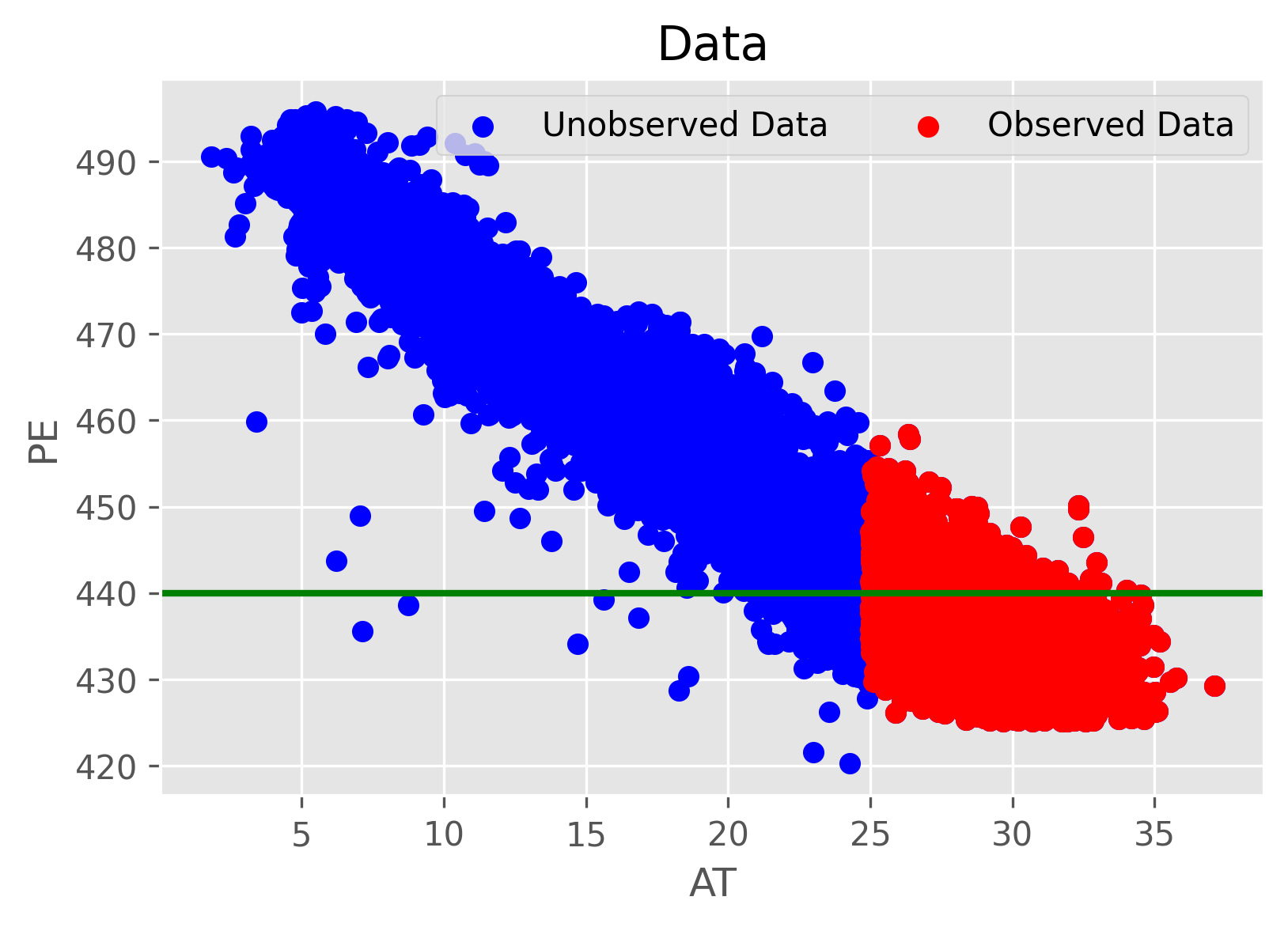}
\caption{Scatterplot of $AT$ and $PE$. The red points show the observed data (used for the model fitting) and the blue points the unobserved data (used for the testing). The green horizontal line indicates where the class changes}
\label{fig:pe_data}
\end{figure}

The data are derived from a combined cycle power plant, described in \cite{tufekci2014prediction}. They consist of four features: the ambient temperature $AT$, the ambient pressure $AP$, the humidity $RH$ and the vacuum $V$, and one output: the electrical energy $PE$. We use the number $465$ as a cut-off point of the $PE$ to create our label $PE_{class}$, as shown by the green horizontal line in Figure~\ref{fig:pe_data}. Therefore, points where $PE \geq 440$ are given the value $PE_{class}=0$, while points where $PE < 440$ are assigned the value $PE_{class}=1$. Our training data are a sample of points where $AT \geq 25^oC$, which is equivalent to collecting data during high temperature days (e.g. summer season). In total the training dataset produced consists of $704$ points for class 0 and $2065$ points for the class 1.  We evaluate the models in the remaining dataset, which consists of $6416$ points for class 0 and $383$ points for the class 1. Note that this is a significantly imbalanced dataset, which adds to the complexity of the problem. Parts of the analysis that follows uses popular techniques to tackle this issue directly.

\subsubsection{Bayesian multivariate logistic regression}

We use a multivariate logistic regression model with parameters the coefficients of the features:
\begin{multline*}
    PE_{class} = \sigma(AT_{co}*AT + AP_{co}*AP + RH_{co}*RH + \\
    V_{co}*V + b) + \epsilon,
\end{multline*}
where $PE_class$ is the response, $AT_{co}$, $AP_{co}$, $RH_{co}$, and $V_{co}$ are the features coefficients described in the previous section, $b$ is the intercept, $\sigma$ is the sigmoid function and $\epsilon$ is Gaussian error with:
\begin{equation*}
    \epsilon \sim \mathcal{N}(0,\sigma^2).
\end{equation*}

Similarly to the previous example, we choose Gaussian distributions for the regression coefficients and intercept, and Exponential for the standard deviation of the Gaussian error:
\begin{align*}
    AT_{co}, AP_{co}, \dots   &\sim \mathcal{N}(0, 10^2), \\
    b & \sim \mathcal{N}(0, 20^2),\\
    \sigma & \sim \operatorname{Exp}(1).
\end{align*}

Evaluation metrics for all the models are shown in Table~\ref{pe_metrics} and Table~\ref{pe_metrics_balance}. In the latter, we repeat the same analyses after upsampling the minority class in the training dataset.

\begin{table}
\caption{Evaluation metrics for the different models.}
\centering
\begin{tabular}{|c||c|c|c|c|}
\hline

Metric/Model & No rules   & $AT$ rules \\ \hline
        Accuracy             & \num{0.9597}     & \num{0.8857} \\ \hline
        AUC              & \num{0.8264}     & \num{0.8805}   \\ \hline
        Sensitivity              & \num{0.6762}     & \num{0.8746} \\ \hline
\end{tabular}
\label{pe_metrics}
\end{table}

\begin{table}
\caption{Evaluation metrics for the different models after class balancing.}
\centering
\begin{tabular}{|c||c|c|c|c|}
\hline

Metric/Model & No rules   & $AT$ rules \\ \hline
        Accuracy             & \num{0.9498}     & \num{0.9506} \\ \hline
        AUC              & \num{0.5683}     & \num{0.8007}   \\ \hline
        Sensitivity              & \num{0.1384}     & \num{0.6318} \\ \hline
\end{tabular}
\label{pe_metrics_balance}
\end{table}

\subsubsection{Rules derivation}\label{rulesderiv-classification}

\begin{figure*}
\centering
  \includegraphics[width=0.6\textwidth]{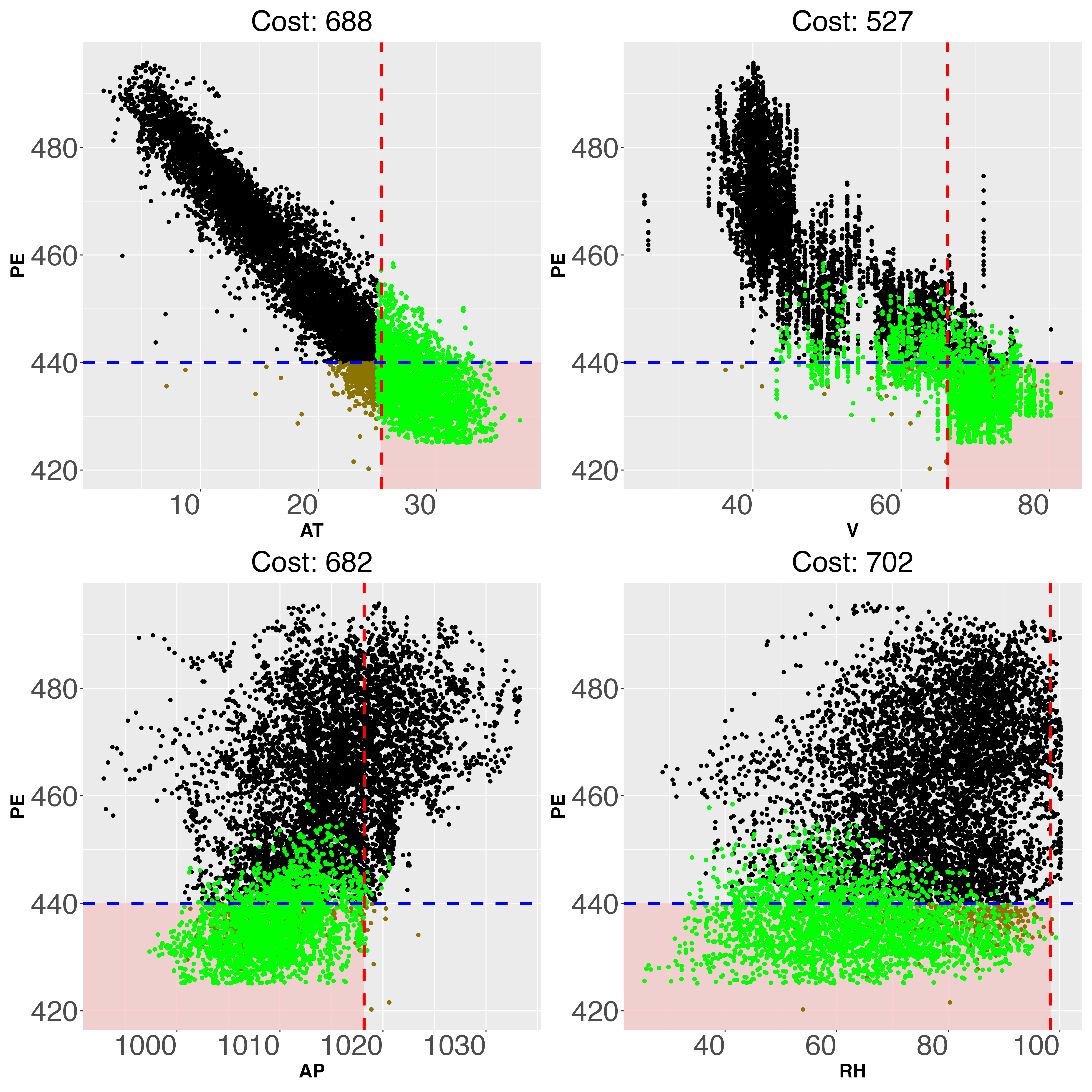}
\caption{Depiction of the rule-bases chosen by grammatical evolution for each feature. The cost on top of each plot indicates the number of training data (in green) that do not follow the rule. The vertical dashed red lines indicate the position of the rules. The horizontal dashed blue line indicates the change of class; Gold points correspond to low and black points to high power output.}
\label{fig:PE_rulesClass}
\end{figure*}

The grammar is presented in Table~\ref{tab:peClass_grammar}.
\begin{table}
\caption{The grammar used for the production of the power plant rules.}
\begin{tabular}{l}
$\mathcal{N} = \{comp, var, num, expr\}$ \\
$\mathcal{T} = \{AT, AP, \dots,>,<,\leq,\geq, -1.4, \dots, 3.5\}$ \\
$\mathcal{S} = <expr>$ \\
\\
$\mathcal{R} = \text{Production rules:} $ \\
$<expr> \quad :== ifelse((comp(var, num)), 1, 0 )$ \\
$<comp> \quad :== >|<|\leq|\geq$ \\
$<var> \quad :== AT|AP|RH|V $ \\
$<num> \quad :== -1.4|-1.35|\dots|3.5 $.
\end{tabular}
\label{tab:peClass_grammar}
\end{table}
Again we need to standardise the data for reasons explained in the previous example. 

The cost function is:
\begin{equation*}
    f(x) = \sum_{i=1}^{N} \left( PE_{class} != y \right),
\end{equation*}
and the optimisation algorithm is evolution strategy.

We examine all the components of $<var>$ sequentially. The results are shown in Figure~\ref{fig:PE_rulesClass}. Note that for visualisation purposes we show the continuous version of the output $PE$, but we use only the categorical version $PE_{class}$ for the grammatical evolution algorithm.

We focus on the rule associated with the $AT-PE_{class}$ pair. We know from \cite{botsas2020rule} and \cite{tufekci2014prediction} that the ambient temperature has some association with the electrical output. After rescaling, the rule is:
\begin{align*}
    R_1&: \text{if} \quad AT < 25.34,\quad \text{then} \quad y = 0,\\
    R_1'&: \text{if} \quad AT \geq 25.34,\quad \text{then} \quad y = 1,
\end{align*}

and the composite rule base ($R_\text{comp}$) is given by
\begin{equation*}
    R_\text{comp} := R_1 \land\ R_i', \quad \text{for} \quad i = 1,2.
\end{equation*}

\subsubsection{Rule-based Bayesian multivariate logistic regression}

We use the rules from the previous section and set $\rules | \pars \sim \operatorname{Beta}(1,1000)$, i.e. a very high level of our confidence in the rules. Sampling is very similar to the one in Algorithm~\ref{alg:sampling steps}. For inspecting whether a point violates the rule we compare $y$ with $0.5$.

The results in Table~\ref{pe_metrics} indicate that the version without the rules performs better in terms of accuracy, whereas the version with the $AT$ rules has a higher sensitivity and is also better in terms of the area-under-curve (AUC) metric. In many real data problems, especially with imbalanced classes, it might be worth applying this trade-off. In this example, it could be important to know when the electrical power output of a plant is very low and adjust our planning accordingly. This problem approach would make sensitivity the most important metric.

\begin{figure}
  \includegraphics[width=\columnwidth]{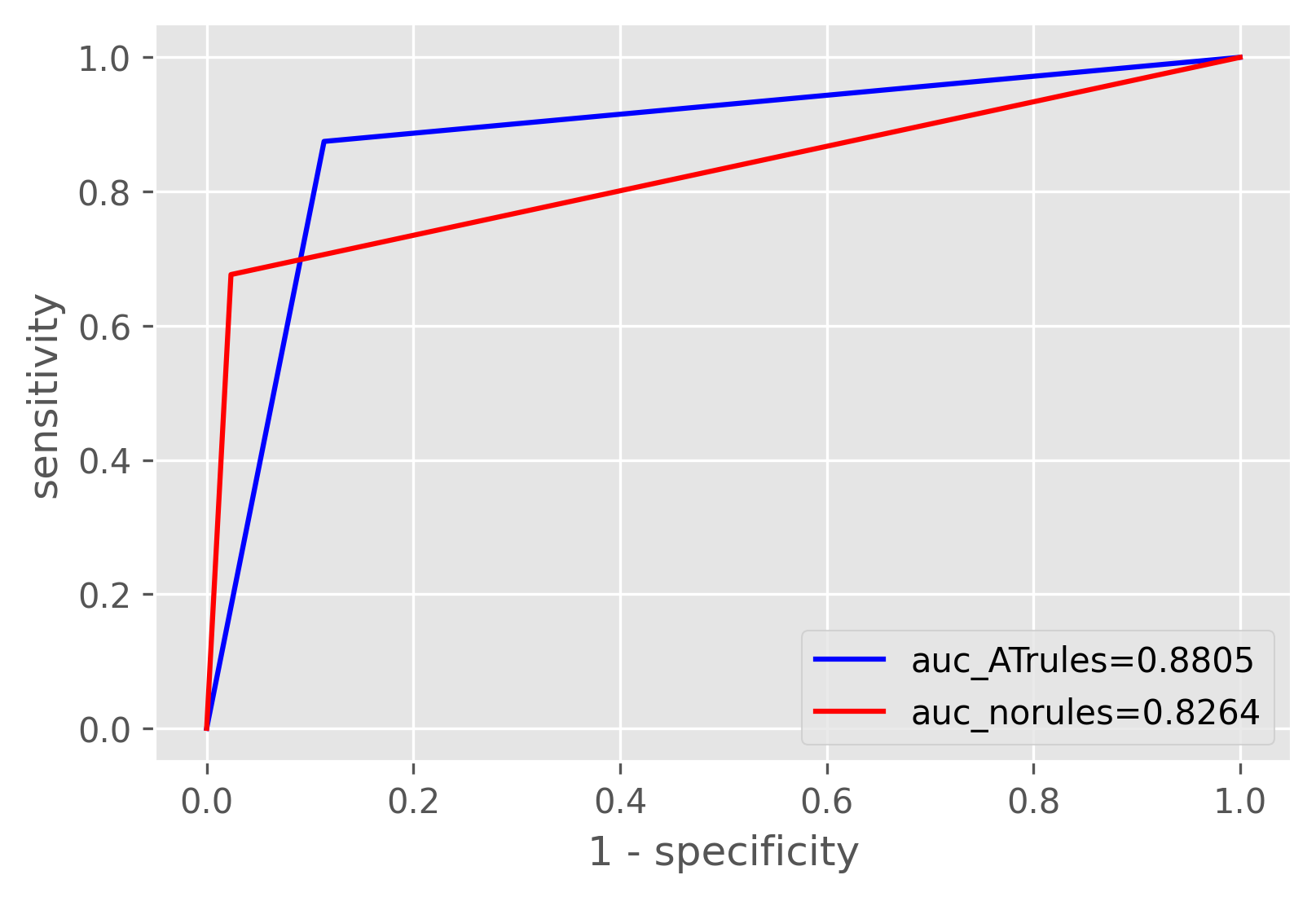}
\caption{ROC curves for the mean posterior prediction of the case without rules (in red) and with $AT$ rules (in blue).}
\label{fig:pe_roccurves}
\end{figure}

The results are confirmed in Figure~\ref{fig:pe_roccurves}, where, for the majority of the plot, the ROC curve that corresponds to the rule version is higher on the y (sensitivity) axis than the no-rule counterpart, which means that for most thresholds the model with the rules will predict more points of class 1 correctly. The area under the ROC curve is also visibly larger in the rule version.

In Table~\ref{pe_metrics_balance} we have repeated the analysis, but with some additional pre-processing in the training set in order to balance the classes. In terms of accuracy the model with the $AT$ rules performs marginally better, while the AUC and sensitivity metrics indicate a significant performance increase for the model with the rules. In almost all metrics, though, we see worse performance than the corresponding results without the balancing (with the accuracy of the model with the $AT$ rules being the only exception). We contribute this to the fact that the imbalance in the training set (where there are more data-points with low electrical output) is different than the imbalance in the testing set (where there are more data-points with high electrical output), and therefore balancing the former was not beneficial.

\subsubsection{Remarks}
In the final example we examined how the methodology can be applied to classification problems. We also saw that the result with the rule versions might not be beneficial in terms of specific metrics (e.g. accuracy), but could be in others (e.g. specificity).

\section{Discussion}\label{discussion}

With the applications of the previous section, we illustrated how the methodology can help derive better results and/or reduce the system uncertainty with the help of automatically derived rules and an appropriate Bayesian context. What is very apparent, though, is that one needs to be very careful when using that methodology. Grammatical evolution attempts to find patterns and associations amongst the data that might not always be meaningful or helpful. This is why this method is better used when combined with (even weak) expert knowledge or domain information, for which there is no real substitute. 

Additionally, there is a question concerning complexity, i.e. how flexible the grammar should be and, consequently, how convoluted the derived rules can be, and whether the final patterns are so intricate that are no longer worth the effort. In general, we believe that it is better to try to derive rules that can reflect something meaningful for the parameter associations, i.e. can be connected or even help discover intuitions, and adjust the grammar accordingly.

We also need to re-iterate limitations associated with the rule-based Bayesian regression context with or without the addition of grammatical evolution. These include the computational complexity linked with the sampling technique, and the potentially complex shape of the rule-based posterior.

Regarding the translation of the methodology into code, we opted for a two-step process. Specifically, we used the gramEvol package from R \citep{noorian2016gramevol} to derive the rules, and the PyMC3 Python package \citep{Salvatier2016} to construct the Bayesian framework. The modular nature and simplicity of the methodology indicate that the requirements to implement it include any grammatical evolution package like PonyGE2 \cite{fenton2017ponyge2}, and any probabilistic framework, such as Stan \citep{stan_cite}, or TensorFlow Probability \citep{abadi2016tensorflow}.

\section{Conclusion}\label{conclusion}

In this paper, we extended our rule-based Bayesian methodology of \cite{botsas2020rule} by introducing a grammatical evolution step, which automates the rule discovery.
We presented the general framework and used the methodology in four applications, adopting different statistical models. In the first application, we derived data from a linear model and we used a uni-variate linear regression model, in the second, we used data from a one-dimensional velocity advection equation and we fit third-degree B-splines, in the third, we used multivariate linear regression models to predict the $CO$ emissions from a gas turbine, and, finally, in the third we used multivariate logistic regression models to predict whether the electrical output of a power plant was high or low.

We extended the rule-based Bayesian regression framework with different variations of the penalty and associated distribution. Other than the proportion of the rule-inputs that violate the rules, modeled with a Beta distribution, that was introduced in \cite{botsas2020rule}, we proposed a penalty based on the total distance from the rule-boundary, modeled with an Exponential distribution and a piece-wise regression penalty with an associated Gaussian distribution. We also presented how we can use the methodology to perform a classification task.

Future research should be focused in applying the methodology to more complex real data applications, where the challenges mentioned in Section \ref{discussion} might be more prominent, such as computational issues and difficulty of assessing the performance of a grammatical evolution-derived rule.

\begin{acknowledgements}
This work was supported by Wave 1 of The UKRI Strategic Priorities Fund under the EPSRC Grant EP/T001569/1, particularly the \emph{Digital Twins for Complex Engineering Systems} theme within that grant and The Alan Turing Institute. IP was partially supported by the NUAcT fellowship at Newcastle University.
\end{acknowledgements}

\section*{Conflict of interest}

The authors declare that they have no conflict of interest.

\bibliography{bibliography}

\end{document}